%% file: main.tex
\documentclass[11pt, a4paper, logo]{eai}
\usepackage{shlab}

\usepackage[authoryear, sort&compress, round]{natbib}
\usepackage[]{mdframed}

\usepackage{dblfloatfix}
\input{math_commands.tex}

\usepackage{listings}
\usepackage{hyperref}
\usepackage{url}
\usepackage{graphicx}
\usepackage{amsmath} %
\usepackage{mathrsfs} %
\usepackage{etoolbox}
\usepackage{cleveref}
\usepackage{tcolorbox}
\usepackage{colortbl}
\usepackage{booktabs}       %
\usepackage{amsfonts}       %
\usepackage{nicefrac}       %
\usepackage{microtype}      %
\usepackage{caption}
\usepackage{subcaption}
\usepackage{algorithm}
\usepackage{algorithmic}
\usepackage{multirow}
\usepackage{lipsum}
\usepackage{makecell} %%%
\usepackage{pifont}  %%%
\usepackage{authblk}  %%%
\usepackage{tikz}
\usepackage[utf8]{inputenc} % allow utf-8 input
\usepackage[T1]{fontenc}    % use 8-bit T1 fonts
\usepackage{hyperref}       % hyperlinks
\usepackage{url}            % simple URL typesetting
\usepackage{booktabs}       % professional-quality tables
\usepackage{amsfonts}       % blackboard math symbols
\usepackage{nicefrac}       % compact symbols for 1/2, etc.
\usepackage{microtype}      % microtypography
\usepackage{xcolor}         % colors
\usepackage{amsfonts}       % blackboard math symbols
\usepackage{multirow}        % multi-row cells in tables
\usepackage{amsmath}
\usepackage{makecell}
\usepackage{pifont} 
\usepackage{wrapfig}

\usepackage{placeins}

\usepackage{booktabs} %
\usepackage{enumitem}%
\setlist[itemize]{noitemsep, topsep=0pt}
\usepackage{enumitem,kantlipsum}

\newcommand{\code}[2]{%
  \begin{tabular}[c]{@{}c@{}}%
    \includegraphics[height=1.5em,width=2.60em,keepaspectratio]{#1}\\[-14pt]%
    {\scriptsize #2}%
  \end{tabular}%
}

\newlength\savewidth

\definecolor{baselinecolor}{HTML}{d6eaf8}

\definecolor{mygray}{gray}{0.4}

\AtBeginEnvironment{tcolorbox}{\tiny}

\newcount\Comments  %
\usepackage{color}
\definecolor{darkred}{rgb}{0.9,0,0}
\definecolor{darkgreen}{rgb}{0,0.5,0}
\definecolor{darkblue}{rgb}{0,0,0.7}
\definecolor{purple}{rgb}{.6, 0,.6}
\definecolor{orange}{rgb}{1.0,0.64,0}
\definecolor{gptgray}{RGB}{245,246,248}
\definecolor{gptbar}{RGB}{230,232,236}
\definecolor{qwenlav}{RGB}{244,240,252}
\definecolor{qwenbar}{RGB}{228,220,246}
\definecolor{glmblue}{RGB}{235,242,250}  % very light blue
\definecolor{catalogblue}{RGB}{19,28,165}
\definecolor{catalogbg}{RGB}{247,248,255}
\definecolor{appendixred}{RGB}{220,35,35}
\newcommand{\kibitz}[2]{\ifnum\Comments=1\textcolor{#1}{#2}\fi}

\title{SetGRPO: Calibrating Grader-Induced Policy Optimization}

\renewcommand{\today}{2026-10-27}
\makeatletter

\newcommand{\Envelope}{%
  \tikz[
    baseline=-0.5ex,
    x=1.15ex,
    y=0.78ex,
    line width=0.32pt
  ]{%
    \draw[draw=black,fill=white] (0,0) rectangle (1,1);
    \draw[draw=black] (0,1) -- (0.5,0.42) -- (1,1);
  }%
}

\def\@author{%
  \begin{tabular}{c}
    Jiale Huang\textsuperscript{1} \quad
    Zixu Li\textsuperscript{1,}\textsuperscript{\textdagger} \quad
    Zhiwei Chen\textsuperscript{1} \quad
    Zhiheng Fu\textsuperscript{1} \quad
    Jinhe Bi\textsuperscript{2} \quad
    Yupeng Hu\textsuperscript{1,}\textsuperscript{\Envelope} \\
    \textsuperscript{1}Shandong University \qquad
    \textsuperscript{2}National University of Singapore \quad
  \end{tabular}%
  \\
\texttt{\footnotesize\color{black!80} 
huangjiale.cs@outlook.com,~\{lizixu.cs,~zivczw,~fuzhiheng8\}@gmail.com}
\\
\texttt{\footnotesize\color{black!80}
bijinhe@outlook.com,~huyupeng@sdu.edu.cn
}
\\
\code{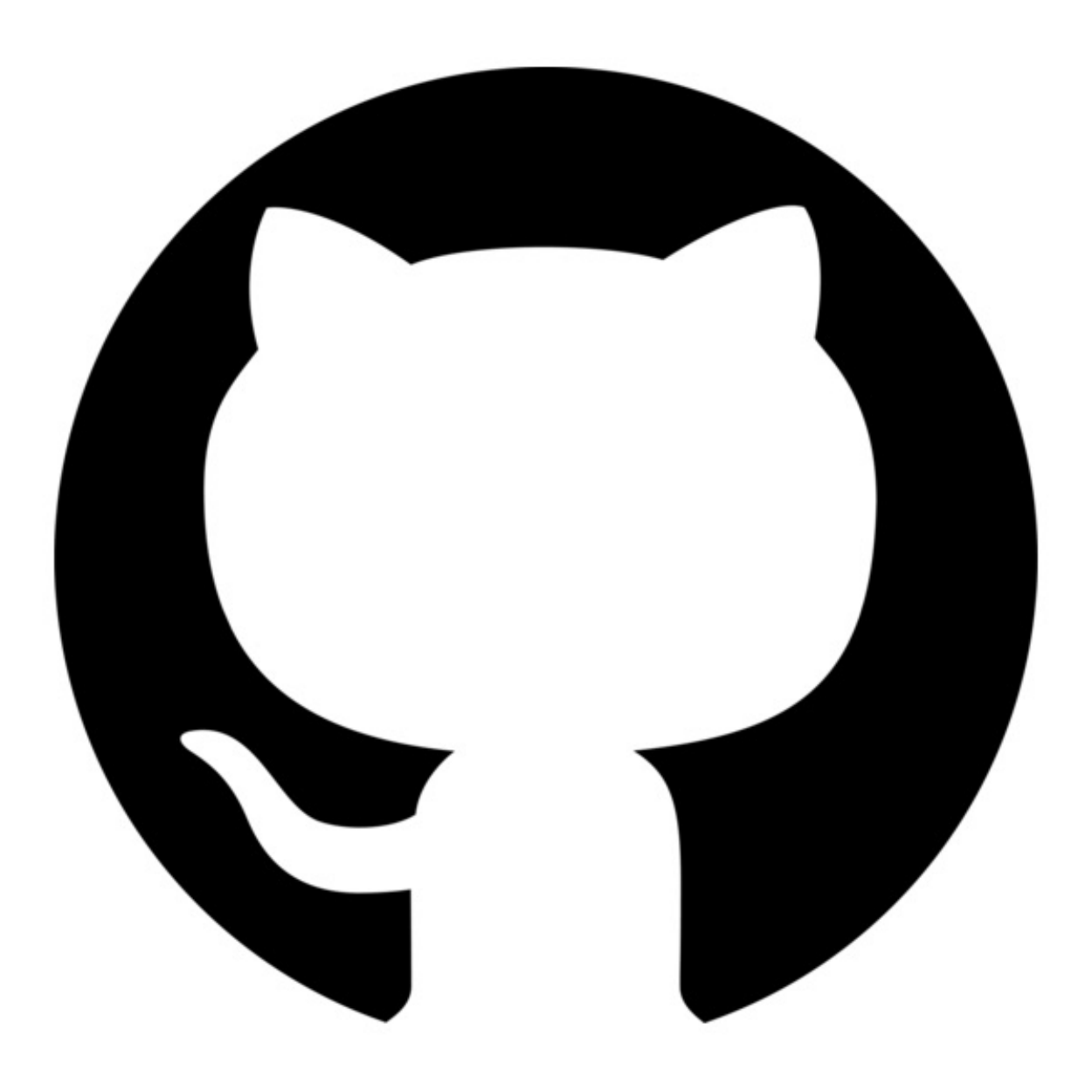}{}\texttt{\footnotesize~Code:~\url{https://github.com/ArcadiaDream/SetGRPO}}
}

\renewcommand{\footnoterule}{%
  \kern -3pt
  \hrule width \textwidth height 0.4pt
  \kern 2.6pt
}

\providecommand{\footrule}{}

\renewcommand{\footrule}{}

\makeatother

\begin{document}

\begin{abstract}
\input{sec/0_abstract} %\input{sections/abstract}
\end{abstract}

\maketitle

\begingroup
\renewcommand{\thefootnote}{}
\footnotetext{%
    † Project Leader: Zixu Li. \quad
  \Envelope\,Corresponding Author: Yupeng Hu.\\[-2pt]
}
\endgroup

% \begin{figure}[h!] %[htbp]
%     \centering
%     \includegraphics[width=0.98\linewidth]{images/internscenes_teaser.png}
%     \vspace{-6pt}
%     \caption{InternScenes is a large-scale, simulatable indoor scene dataset with diverse layouts and various 3D objects. It supports various tasks, such as scene layout generation and vision navigation.}
%     \label{Teaser}
%     \vspace{-8pt}
% \end{figure}

\input{sec/1_intro}
\input{sec/2_related}
\input{sec/3_preliminary}

\input{sec/4_method}
% \input{sec/supplement_3_dataset_show_details}
\input{sec/5_experiment}
\input{sec/6_conclusion}

\bibliography{main}

% \appendix
% \input{sec/appendix_1_proof}
% \input{sec/appendix_2_related}
% \input{sec/appendix_3_implementation_details}
% \input{sec/appendix_4_additional_quantitative}
% \input{sec/appendix_5_algorithm}
% \input{sec/appendix_6_additional_qualitative}
% \newpage
% \input{sec/supplement_1_dataset_building_details}
% \input{sec/supplement_2_experiment_details}

% \input{sec/supplement_5_effi}
% \input{sec/supplement_7_discuss_infinigen}
% \input{sec/supplement_6_articulated_surport}

\end{document}

%% file: math_commands.tex
\usepackage{amsmath,amsfonts,bm}
\usepackage{multirow}
\usepackage{subcaption}
\usepackage{wrapfig}

\def\eqref#1{equation~\ref{#1}}

\def\1{\bm{1}}

\DeclareMathAlphabet{\mathsfit}{\encodingdefault}{\sfdefault}{m}{sl}
\SetMathAlphabet{\mathsfit}{bold}{\encodingdefault}{\sfdefault}{bx}{n}

%% file: sec/0_abstract.tex
Rubric-based reinforcement learning uses fine-grained grader feedback to train large language models on open-ended tasks, but grader randomness can alter GRPO updates in ways that score discrepancies alone cannot characterize. Because GRPO jointly normalizes rewards within each response group and combines the resulting training coefficients with response-level gradients, a scoring discrepancy can shift the coefficients of the entire group and induce policy-dependent update changes. Selective execution further affects the direction-reversal risk of retained updates.
To address this, we propose \textbf{Set}-Valued \textbf{G}roup \textbf{R}elative \textbf{P}olicy \textbf{O}ptimization (\textbf{SetGRPO}). It comprises three modules: \textit{(a) Joint Reward Set Calibration} constructs a joint uncertainty set for the training coefficients; \textit{(b) Geometry-Guided Safe Recovery} maps candidate coefficients into the current policy's local update geometry and determines the largest safe recovery ratio for each response group; and \textit{(c) Selective Risk-Gated Optimization} selects a global threshold through independent risk calibration to control the conditional direction-reversal risk of accepted updates while preserving the original GRPO objective. Under exchangeability, SetGRPO provides finite-sample marginal coverage for the joint coefficient set and simultaneous risk control across candidate thresholds. On ten open-ended benchmarks spanning three domains, SetGRPO achieves the highest macro average among all compared training methods on both Qwen3.5 and Llama3.2 model families. 
% The code is available at \href{https://anonymous.4open.science/r/SetGRPO-73EB}{https://anonymous.4open.science/r/SetGRPO-73EB}.

% \links{
%   \link{github}{Code}{https://github.com/InternRobotics/InternScenes}, 
%   \link{huggingface}{Data}{https://huggingface.co/datasets/InternRobotics/InternScenes}, 
%   \link{homepage}{Homepage}{https://marjordcpz.github.io/InternScenes.github.io/}, 
%   % \link{demo}{https://demo.com}, 
%   % \link{paper}{https://arxiv.org/abs/1234.5678}
% }

%% file: sec/1_intro.tex
% !TEX root = ../neurips_2025.tex
\section{Introduction}
\label{sec:introduction}

% intro

Nowadays, Rubric-based Reinforcement Learning (Rubric-based RL) has emerged as an important training paradigm for improving the performance of large language models on open-ended generation tasks~\citep{li2026rubrichub,zhou2025ruscarl,kim2023prometheus,ye2023flask}. It uses fine-grained rubrics and LLM judges to evaluate open-ended responses across dimensions such as relevance, completeness, factuality, and style~\citep{li2026rubrichub,zhou2025ruscarl,kim2023prometheus,ye2023flask,liu2023geval,zheng2023judging}. 
These dimension-level scores are aggregated into response-level rewards that
serve as training signals for reinforcement learning
~\citep{ouyang2022training,stiennon2020learning}, thereby accommodating
grader's score discrepancies during training and enabling optimization of
open-ended generation tasks. Related work has also explored multimodal fusion,
video-language adaptation, and parameter-efficient training for large models
~\citep{song3,song5,li2026lora}.

% It organizes evaluation with fine-grained rubrics~\citep{kim2023prometheus,ye2023flask}, which assess response quality across multiple dimensions including relevance, completeness, factuality, and expressive quality~\citep{liu2023geval,ye2023flask}. With the aid of LLM judge~\citep{zheng2023judging,liu2023geval,kim2023prometheus}, the grader then generates a reward for each dimension~\citep{liu2023geval,kim2023prometheus}. Such multi-dimensional evaluation mitigates the quality differences that a single overall score would mask~\citep{kim2023prometheus,ye2023flask}. The per-dimension scores are then aggregated into a response-level reward, which serves as the training signal for downstream reinforcement learning~\citep{ouyang2022training,stiennon2020learning}, thereby accommodating grader's score discrepancies during training and enabling optimization of open-ended generation tasks.
% 近年来，为提升大语言模型在开放式生成任务中的表现，Rubric-based Reinforcement Learning (Rubric-based RL) 逐渐成为一种重要的训练范式~\citep{li2026rubrichub,zhou2025ruscarl,kim2023prometheus,ye2023flask}。该范式采用细粒度 rubric 组织评价~\citep{kim2023prometheus,ye2023flask}，which通过相关性、完整性、事实性和表达质量等多个维度评价回答质量~\citep{liu2023geval,ye2023flask}，并借助LLM judge~\citep{zheng2023judging,liu2023geval,kim2023prometheus}，由 grader 依据各维度生成 reward~\citep{liu2023geval,kim2023prometheus}。这种多维度评价能够减少单一总体分数掩盖的质量差异~\citep{kim2023prometheus,ye2023flask}。随后，各维度得分被聚合为回答级 reward，并作为后续强化学习的训练信号~\citep{ouyang2022training,stiennon2020learning}，从而在训练中兼顾 grader 的评分差异，实现开放式生成任务优化。

Existing rubric-based RL studies build on GRPO~\citep{shao2024deepseekmath,deepseekr1} and examine reward over-optimization~\citep{gao2022overoptimization,lambert2024rewardbench}, LLM-judge consistency and bias~\citep{zheng2023judging,wang2023faireval,liu2023geval}, fine-grained evaluation~\citep{kim2023prometheus,ye2023flask,liu2023geval}, rubric constructions and rubric-based training for open-ended tasks~\citep{li2026rubrichub,zhou2025ruscarl}. Recent reinforcement-learning studies have also explored how to improve the utility of sampled trajectories through reflection, trajectory reuse, and rollout transformation~\citep{bi2026reflectrllearninggoldennegative,bi2026echorl}. Other work examines selection and self-evaluation mechanisms in multimodal or embodied settings~\citep{aniri2026fabrimaeitrustmyself,Bi2025PRISMSI,bi-etal-2025-llava}. Related studies further investigate reasoning, logical behavior, and the gap between model-based preference and real user engagement~\citep{zhang2026too,song8,song10}.
However, these studies largely focus on the construction, reuse, or assessment of training signals and do not characterize how stochastic grader scores propagate through GRPO's within-group normalization and gradient aggregation to alter local policy updates under the current gradient geometry. Specifically, for a given response group, GRPO derives the training coefficients from the shared mean and scale of the group rewards and combines them with the response-level gradients of each response under the current policy to form a local policy update. We therefore conjecture that \textit{a change in a single reward alters multiple training coefficients within the group through the shared statistics, and that different policy snapshots induce different update outcomes}.

To verify this conjecture, we score fixed response groups twice with the same grader and compare the two scoring results pairwise (detailed in Section~\ref{sec:preliminaries}). As shown in Figure~\ref{fig:intro_joint_diagnostics}(a), we first compare score discrepancies with the coefficient residuals obtained after within-group normalization, testing whether shared statistics propagate score discrepancies to the entire group of coefficients. As shown in Figure~\ref{fig:intro_joint_diagnostics}(b), we further compare the local updates of the same response group under different policy snapshots, testing whether the effect of coefficient discrepancies depends on the gradient geometry of the current policy. The results show that shared normalization propagates score discrepancies to multiple coefficients within a group, whereas similar coefficient shifts can induce different update directions under different policy geometries, thereby changing which groups are retained by selective execution. These findings support our conjecture and motivate the following question: \textit{how can we characterize the effect of the training coefficients used by the optimizer on the current policy update while preserving effective training progress and avoiding directional deviation?}

% The results show that score discrepancies affect multiple training coefficients within a group simultaneously through the shared statistics, and that even when training coefficients are close, changing the policy snapshot affects the update direction and thereby which response groups are retained after selective execution. This finding confirms the conjecture proposed above. Accordingly, we formulate the research problem as follows: \textit{how can we characterize the influence of the within-group training coefficients actually used by the optimizer on the current policy update, while preserving effective training progress and avoiding deviation in update direction?}

% 为验证上述猜想，我们在固定回答组上用相同grader进行两次评分，将评分结果进行两两比较(detailed in 第~\ref{sec:preliminaries}~节)。如图~\ref{fig:intro_joint_diagnostics}(a)所示，我们首先比较评分差异及其经组内归一化后的系数残差，用于检验评分差异是否会被共享统计量传递到整组系数。此外，如图~\ref{fig:intro_joint_diagnostics}(b)所示，我们进一步在不同策略快照下比较相同回答组的局部更新，用于检验系数差异的影响是否依赖当前策略的梯度几何。结果显示，评分差异会通过共享统计量同时影响组内多个训练系数，且即使训练系数相近，策略快照的变化也会影响更新方向，并进一步影响选择性执行后哪些回答组被保留。这一结论验证了我们在前文中所提出的猜想。
% 基于上述结论，我们将研究问题概括为：如何以训练器实际使用的组内训练系数为对象，刻画其对当前策略更新的影响，并在保留有效训练进度的同时避免更新方向偏差？

To address this problem, we propose \textbf{Set}-Valued \textbf{G}roup \textbf{R}elative \textbf{P}olicy \textbf{O}ptimization (\textbf{SetGRPO}), which models GRPO's within-group training coefficients as a joint uncertainty set and uses it to select response groups and construct policy updates. Concretely, SetGRPO comprises three components: (1) \textit{Joint Reward Set Calibration}, which uses paired reward discrepancies and independent calibration data to characterize the joint coefficient shifts induced by within-group normalization; (2) \textit{Geometry-Guided Safe Recovery}, which maps candidate coefficients through the current policy's response-level gradients and determines the largest direction-safe recovery ratio; and (3) \textit{Selective Risk-Gated Optimization}, which selects a global threshold on an independent risk-calibration set to control the conditional direction-reversal risk of accepted updates. During training, only groups satisfying this threshold contribute their recovered coefficients to the original GRPO objective, preserving effective updates while limiting overall update risk.

Our main contributions are summarized as follows:
\begin{itemize}[leftmargin=8pt]
\item We first investigate the propagation mechanism of rubric score discrepancies in GRPO: a score discrepancy couples the training coefficients of an entire group through the shared mean and scale, and then jointly with the response-level gradients of the current policy determines the local update, yielding different update magnitudes and directions.
\item We formalize this propagation process as a training-coefficient-driven policy-update problem and propose SetGRPO, which conformally calibrates GRPO's joint coefficient space and combines policy-update geometry with selective risk gating to limit direction deviation while preserving effective training progress.
\item We systematically evaluate SetGRPO on ten open-ended task benchmarks across three domains with the Qwen3.5 and Llama-3.2 model families. The results show that SetGRPO achieves the best performance among all compared training methods on both model families, confirming its effectiveness in improving the reliability of GRPO training under rubric score discrepancies.
\end{itemize}

% 本文的主要贡献总结如下：
% \begin{itemize}[leftmargin=8pt]
%     \item 我们首次探索了rubric评分差异在 GRPO 中的传播机制，即评分差异会先经由组内共享的均值和尺度耦合整组训练系数，随后与当前策略的 response-level gradients 共同决定局部更新，从而导致不同的更新幅度和方向。

%     \item 我们将上述传播过程形式化为训练系数驱动的策略更新问题，并提出SetGRPO。该方法在 GRPO 实际使用的联合训练系数空间中进行 conformal calibration，再结合策略更新几何和选择性风险门控，控制接受更新的总体方向偏差，同时尽量保留有效训练进度。

%     \item 我们在三个领域的十个开放式任务基准上，基于 Qwen3.5 和 Llama-3.2 模型簇 对 SetGRPO 进行系统评估。结果表明，SetGRPO 在两个模型上均取得所有对比训练方法中的最优性能，验证了其在 rubric 评分差异下提升 GRPO 训练可靠性的有效性。
% \end{itemize}

%% file: sec/2_related.tex
% !TEX root = ../neurips_2025.tex
\vspace{-8pt}
\section{Related Work}
\label{sec:related}

\noindent\textbf{Reinforcement Learning for Large Language Models and Rubric Feedback.}
Alignment training for large language models has evolved from preference learning and supervised fine-tuning to human- or AI-feedback-based instruction alignment~\citep{christiano2017deep,ziegler2019fine,stiennon2020learning,ouyang2022training,bai2022constitutional}, giving rise to PPO and reward-model-free DPO~\citep{schulman2017proximal,rafailov2023direct}. For reasoning tasks, verifiers and process supervision provide fine-grained training signals~\citep{cobbe2021training,uesato2022solving,lightman2023verify},while DeepSeekMath and DeepSeek-R1 apply GRPO to large-language-model reasoning~\citep{shao2024deepseekmath,deepseekr1}. Beyond standard single-agent training, recent work has also studied group-agent interaction and long-horizon social simulation in multi-agent settings~\citep{song12,song14,song2026socialintelligence}.
Related work has also explored composed image retrieval and egocentric
reasoning~\citep{median,OmniEgo-R2}.For open-ended tasks without unique reference answers, G-Eval, FLASK,MT-Bench, and Prometheus use rubrics or LLM judges for structured evaluation~\citep{liu2023geval,ye2023flask,zheng2023judging,kim2023prometheus},while related retrieval studies consider semantic enhancement, focus revision,and multi-modification alignment~\citep{imagine,OFFSET,TEMA}.
RewardBench, RubricHub, and RuscaRL assess reward-model capability or use rubric feedback for reinforcement learning~\citep{lambert2024rewardbench,li2026rubrichub,zhou2025ruscarl}. Despite evidence of reward overoptimization and systematic judge bias~\citep{gao2022overoptimization,wang2023faireval,zheng2023judging}, prior work largely treats grader outputs as fixed training signals and does not trace how the score discrepancies propagate through GRPO's group normalization and response-level gradient aggregation to alter the actual policy update~\citep{bi2026the,aniri2026opdvvisualonpolicyselfdistillation}, while
related robustness-oriented studies in multimodal retrieval address noisy
supervision and distributional variation through noise mitigation and
invariance modeling~\citep{INTENT,rankvr,Air-know,ConeSep}.

\noindent\textbf{Conformal Prediction and Selective Risk Control.}
The geometry of reasoning trajectories has also been studied through layerwise trajectory evolution and self-evaluation~\citep{bi2026the}.
Conformal prediction exploits exchangeability to construct, from calibration samples, prediction sets with finite-sample coverage guarantees~\citep{shafer2008tutorial,angelopoulos2022gentle}, and has extended from regression and classification coverage to input-dependent uncertainty and general risk functions~\citep{lei2018distribution,romano2019conformalized,romano2020classification,bates2021distribution,angelopoulos2022crc}. Selective prediction controls the risk on an accepted subset through abstention or sample screening
~\citep{geifman2019selectivenet,Bi2025PRISMSI}.Beyond this setting, related video and egocentric systems have also studied temporal reranking, scene adaptation, and reliability-aware action composition ~\citep{REFINE,EgoAdapt,habit}.The Clopper--Pearson interval provides a classical one-sided upper confidence bound for finite-sample binomial risk\citep{clopper1934use}. SetGRPO draws on these ideas of calibration and risk control, but shifts the object of calibration to the joint training coefficients actually used by GRPO and the policy updates they induce, and further controls the conditional direction-reversal risk over the set of accepted updates, while related work has also considered calibration and reliability in retrieval and search settings~\citep{ReTrack,LightAIR,STABLE}.

% Beyond language-model reinforcement learning, robustness- and
% reliability-oriented learning has also been studied in multimodal retrieval
% and vision-language applications, including noise-robust composed retrieval,
% calibration-based retrieval, reliability-aware search, and
% self-evaluation-based adaptation~\citep{ConeSep,Air-know,REFINE,OFFSET,ReTrack,
% INTENT,TEMA,LightAIR,STABLE,OmniEgo-R2,TempRet,EgoAdapt,EgoAction,R3,
% bi-etal-2025-llava,aniri2026opdvvisualonpolicyselfdistillation,
% aniri2026fabrimaeitrustmyself}. These works provide broader evidence that
% robustness, calibration, and selective decision-making are useful design
% principles across learning systems.
% \noindent\textbf{Conformal Prediction 与选择性风险控制。}
% Conformal prediction 利用 exchangeability 从校准样本构造具有有限样本覆盖保证的预测集合\citep{shafer2008tutorial,angelopoulos2022gentle}，并已从回归和分类覆盖扩展到输入相关的不确定性和一般风险函数\citep{lei2018distribution,romano2019conformalized,romano2020classification,bates2021distribution,angelopoulos2022crc}。选择性预测则通过拒答或样本筛选控制接受子集上的风险\citep{geifman2019selectivenet}，Clopper--Pearson 区间为有限样本二项风险提供了经典的单侧上置信界\citep{clopper1934use}。SetGRPO 借鉴这些校准与风险控制思想，但将校准对象转移到 GRPO 实际使用的联合训练系数及其诱导的策略更新，并进一步控制接受更新集合上的条件方向反转风险。

%% file: sec/3_preliminary.tex
% !TEX root = ../neurips_2025.tex
\vspace{-6pt}
\section{Preliminary}
\vspace{-6pt}

\label{sec:preliminaries}

In rubric-based reinforcement learning, grader rewards are normalized within each GRPO response group to produce training coefficients, which are combined with the current policy's response-level gradients to form local updates. We therefore analyze score discrepancies along this reward-to-update chain, first formalizing the mapping and then evaluating its training consequences through paired pilot experiments.

% the response-level rewards produced by the grader first undergo GRPO's within-group normalization to yield training coefficients, which are then combined with the response-level gradients of the current policy over each response to form a local update. The effect of grader score discrepancies on training must therefore be analyzed along the computational chain of rewards, within-group normalization, and policy update. In this section, we first formally define this mapping and then examine its consequences for training through paired pilot experiments.

% 在 rubric-based reinforcement learning 中，grader评分产生的回答级 reward 先经 GRPO 的组内归一化得到训练系数，再与当前策略在各回答上的 response-level gradients 结合形成局部更新。因此，grader的评分差异对训练的影响需要沿着 reward、组内归一化和策略更新这一计算链条进行分析。在这一章节，我们先形式化定义这一映射，再通过配对预实验考察其训练后果。

\subsection{Rubric-based GRPO}
\label{subsec:grpo}

% 对于问题根 $x$，rollout policy $\pi_{\boldsymbol\theta_{\mathrm{old}}}$ 生成包含 $K$ 个回答grader根据给定 rubric 产生回答级 reward。记回答组、奖励向量及原始 GRPO 训练系数为
% \begin{equation}
%     \mathbf Y_x=(y_{x,1},\ldots,y_{x,K}),
%     \qquad
%     \mathbf r_x=(r_{x,1},\ldots,r_{x,K})^\top,
%     \qquad
%     \hat{\mathbf a}_x=T_{\mathrm{train}}(\mathbf r_x),
%     \label{eq:nominal_coefficient}
% \end{equation}
% 其中，$T_{\mathrm{train}}$ 表示原始 GRPO 对一组 reward 进行组中心化和尺度归一化的训练映射，$\hat{\mathbf a}_x$ 为grader评分得到的 reward产生的训练系数。由于组内回答共享中心和尺度，每个 $\hat a_{x,i}$ 都由整组 reward 共同决定。

% 随后，这些系数作为回答级 advantage 进入原始 GRPO 目标，定义组级目标为
% \begin{equation}
%     \mathcal J_{\mathrm{GRPO}}(\boldsymbol\theta;\hat{\mathbf a}_x)=
%     \mathbb E_x\left[\frac{1}{K}\sum_{i=1}^{K}\frac{1}{|y_{x,i}|}
%     \sum_{t=1}^{|y_{x,i}|}\mathcal L_{\mathrm{GRPO}}\left(
%     \omega_{x,i,t}(\boldsymbol\theta),\hat a_{x,i}\right)\right],
% \label{eq:grpo_objective}
% \end{equation}
% 其中，
% $\omega_{x,i,t}(\boldsymbol\theta)=
% \frac{\pi_{\boldsymbol\theta}(y_{x,i,t}\mid x,y_{x,i,<t})}
% {\pi_{\boldsymbol\theta_{\mathrm{old}}}(y_{x,i,t}\mid x,y_{x,i,<t})},$为token-level probability ratio；$\mathcal L_{\mathrm{GRPO}}(\omega,a)$ 表示保留 probability ratio 与 clipping mechanism 的 token-level surrogate term。

For a problem $x$, the rollout policy $\pi_{\boldsymbol\theta_{\mathrm{old}}}$ generates a group of $K$ responses, over which the grader produces response-level rewards according to the given rubric. Denote the response group, the reward vector, and the nominal training coefficients of GRPO as
\begin{equation}
\small
    \mathbf Y_x=(y_{x,1},\ldots,y_{x,K}),
    \qquad
    \mathbf r_x=(r_{x,1},\ldots,r_{x,K})^\top,
    \qquad
    \hat{\mathbf a}_x=T_{\mathrm{train}}(\mathbf r_x),
    \label{eq:nominal_coefficient}
\end{equation}
where $T_{\mathrm{train}}$ denotes the training mapping of the original GRPO that group-centers and scale-normalizes a group of rewards, and $\hat{\mathbf a}_x$ denotes the training coefficients induced by the rewards produced by the grader. Because the responses within a group share the same center and scale, each $\hat a_{x,i}$ is determined jointly by the rewards of the entire group.

These coefficients then enter the original GRPO objective as response-level advantages. We define the group-level objective as
\begin{equation}
\small
    \mathcal J_{\mathrm{GRPO}}(\boldsymbol\theta;\hat{\mathbf a}_x)=
    \mathbb E_x\left[\frac{1}{K}\sum_{i=1}^{K}\frac{1}{|y_{x,i}|}
    \sum_{t=1}^{|y_{x,i}|}\mathcal L_{\mathrm{GRPO}}\left(
    \omega_{x,i,t}(\boldsymbol\theta),\hat a_{x,i}\right)\right],
\label{eq:grpo_objective}
\end{equation}
where $\omega_{x,i,t}(\boldsymbol\theta)=
\frac{\pi_{\boldsymbol\theta}(y_{x,i,t}\mid x,y_{x,i,<t})}
{\pi_{\boldsymbol\theta_{\mathrm{old}}}(y_{x,i,t}\mid x,y_{x,i,<t})}$ is the token-level probability ratio, and $\mathcal L_{\mathrm{GRPO}}(\omega,a)$ denotes the token-level surrogate term retaining the probability ratio and the clipping mechanism.

These training coefficients determine each response's relative weight in the group objective but not the update direction, which also depends on the response-specific gradient directions and their mutual relations under the current policy. Let $\mathbf g_{x,i}$ denote the mean response log-probability gradient for $y_{x,i}$. Performing a first-order expansion around the current rollout policy and neglecting the learning rate and the common scaling across groups, the contribution of the coefficient vector $\mathbf a$ to the local update of the response group is
% These training coefficients determine the relative contribution of each response in the group-level objective, but they do not by themselves determine the direction of the parameter update. For different responses, the same coefficient is multiplied by a different local gradient, so the final update depends on both the magnitude of the coefficients and the directions of the response gradients under the current policy and their mutual relations. Denote the mean response log-probability gradient of response $y_{x,i}$ as $\mathbf g_{x,i}$. Performing a first-order expansion around the current rollout policy and neglecting the learning rate and the common scaling across groups, the contribution of the coefficient vector $\mathbf a$ to the local update of the response group is
% 这些训练系数决定各回答在组级目标中的相对贡献，但它们本身还不能决定参数更新的方向。对于不同回答，同样的系数会与不同的局部梯度相乘，因此，最终更新既取决于系数的大小，也取决于当前策略下各回答梯度的方向及其相互关系。记回答 $y_{x,i}$ 的平均 response log-probability gradient 为 $\mathbf g_{x,i}$。在当前 rollout policy 附近进行一阶展开，并忽略学习率和组间公共缩放后，系数向量 $\mathbf a$ 对回答组局部更新的贡献为，
\begin{equation}
\small
\begin{aligned}
   \mathbf g_{x,i}&=\nabla_{\boldsymbol\theta}
   \left[\frac{1}{|y_{x,i}|}\sum_{t=1}^{|y_{x,i}|}
   \log\pi_{\boldsymbol\theta}\left(y_{x,i,t}\mid x,y_{x,i,<t}\right)\right],&
   \Delta\boldsymbol\theta_x(\mathbf a)
   \propto\sum_{i=1}^{K}a_i\mathbf g_{x,i}.
\end{aligned}
\label{eq:local_update_preliminary}
\end{equation}

\subsection{Preliminary Study on Grader-Induced Update Risk}
\label{subsec:preliminary_study}

The above equation reveals the propagation chain: \textit{a score discrepancy changes the group's training coefficients, which modifies the local update through the current policy's gradients and ultimately affects which response groups are retained.} To assess this risk and motivate SetGRPO, we conduct pilot experiments asking whether (i) within-group normalization couples scoring errors, (ii) the effect of coefficient shifts depends on the policy state, and (iii) selective execution changes the aggregate direction-reversal risk.
% The above equation specifies the concrete path through which score discrepancies affect the policy update: \textit{a score first changes the training coefficients of a group, then alters the update through the local gradients of the current policy, and ultimately determines which response groups are retained.} To test whether this path poses a practical risk and to provide the basis for SetGRPO, we conduct pilot experiments around three questions: whether within-group normalization couples scoring errors, whether the update consequences of a coefficient shift depend on the current policy, and whether selective execution changes the aggregate direction-reversal risk.
% 上式给出了评分差异影响策略更新的具体路径：评分先改变组内训练系数，再通过当前策略下的局部梯度改变更新，最终影响哪些回答组会被保留。检验这一路径是否产生实际风险，并为后文的联合奖励集合校准（JRSC）、几何引导的安全恢复（GGSR）和选择性风险门控优化（SRGO）提供依据，我们围绕三个问题开展预实验：组内归一化是否使评分误差相互耦合；系数偏移的更新后果是否依赖当前策略；以及选择性执行是否改变总体方向反转风险。

% ===== Condensed preliminary figure =====
\begin{figure*}[t]
    \centering
    \begin{subfigure}[t]{0.49\textwidth}
        \centering
        \includegraphics[width=\linewidth]{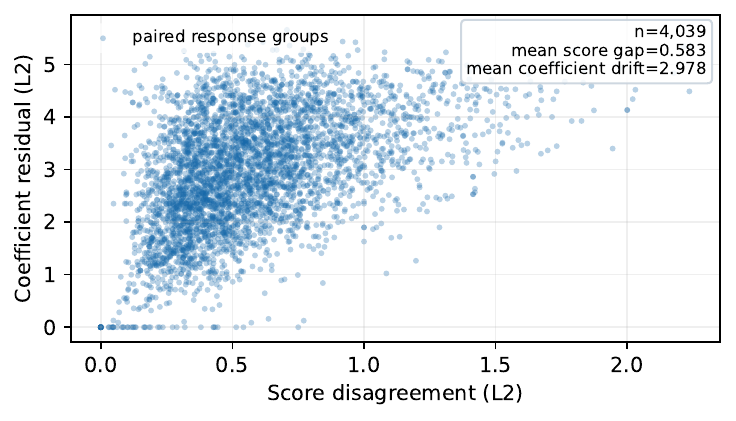}
        \vspace{-20pt}
        \caption{Score disagreement propagates jointly.}
        \label{fig:intro_error_drift}
    \end{subfigure}
    \hfill
    \begin{subfigure}[t]{0.49\textwidth}
        \centering
        \includegraphics[width=\linewidth]{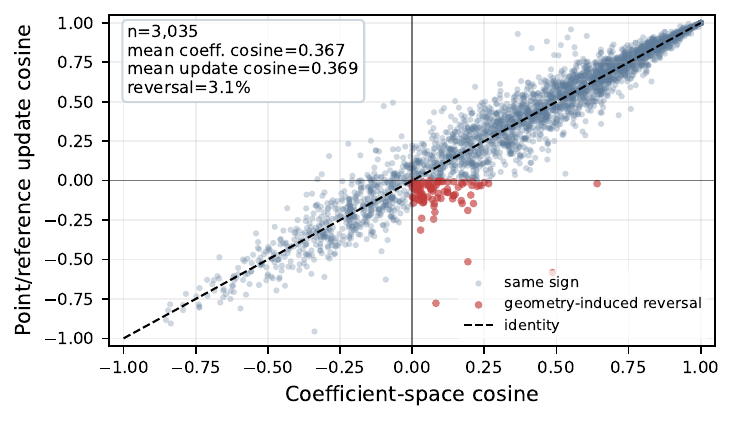}
        \vspace{-20pt}
        \caption{Policy geometry changes local update alignment.}
        \label{fig:intro_update_alignment}
    \end{subfigure}
            \vspace{-8pt}
    \caption{Illustration of score disagreement and its effect on local policy-update geometry.}
            \vspace{-18pt}
    \label{fig:intro_joint_diagnostics}
\end{figure*}

%
% ===== Condensed paired setup =====
\textbf{Setup.}
We run original GRPO on Qwen3.5-9B with the RubricHub training set and collect response groups with their corresponding policy snapshots; training details are given in Section~\ref{sec:experiments}. For each group, we freeze the responses and their order, score it twice with the same grader, and map both scores through $T_{\mathrm{train}}$. For the second scoring, we define
$\mathbf r_x^\star,\quad
\mathbf a_x^\star=T_{\mathrm{train}}(\mathbf r_x^\star),\quad
\boldsymbol\delta_x=\mathbf a_x^\star-\hat{\mathbf a}_x.$
Using the same policy snapshot, we compare the updates induced by the two coefficient sets and analyze coefficient-residual dependence, policy-state dependence, and aggregate risk after group-wise screening.
% We run the original GRPO on Qwen3.5-9B with the Rubrichub training set and collect response groups together with their corresponding policy snapshots; the training configuration is given in Section~\ref{sec:experiments}. For each response group, we freeze the response contents and their internal ordering, score the same group twice with the grader, and map the two scorings into coefficient space through the same $T_{\mathrm{train}}$. We denote the rewards from the second scoring, the corresponding training coefficients, and the paired coefficient residual as
% $\mathbf r_x^\star,\quad \mathbf a_x^\star=T_{\mathrm{train}}(\mathbf r_x^\star),\quad
% \boldsymbol\delta_x=\mathbf a_x^\star-\hat{\mathbf a}_x.$
% For each paired response group, we compare the local updates induced by the two coefficient sets under the same policy snapshot, and analyze the within-group structure of the coefficient residual, the dependence of the local update on the policy state, and the aggregate risk after per-group screening.
% 我们在 Qwen3.5-9B 和 Rubrichub 训练集上运行原始 GRPO，收集回答组及相应策略快照，训练配置见 Section~\ref{sec:experiments}。对每个回答组，我们冻结回答内容及内部顺序，由两次 grader 评分分别对同一组回答评分，并通过相同的 $T_{\mathrm{train}}$ 映射到系数空间。记第二次评分得到的 reward、第二次评分对应的训练系数和配对系数残差为
% $\mathbf r_x^\star,\quad \mathbf a_x^\star=T_{\mathrm{train}}(\mathbf r_x^\star),\quad
% \boldsymbol\delta_x=\mathbf a_x^\star-\hat{\mathbf a}_x.$
% 对于每个配对回答组，我们在相同策略快照下比较两组系数诱导的局部更新，据此分析系数残差的组内结构、局部更新对策略状态的依赖性，以及单组筛选后的总体风险。

\textbf{Does Grader Discrepancy Remain Response-wise Independent?}
By Eq.~\ref{eq:nominal_coefficient} and Eq.~\ref{eq:grpo_objective}, group normalization uses a shared center and scale. Therefore, as shown in Figure~\ref{fig:intro_joint_diagnostics}(a), since $T_{\mathrm{train}}$ normalizes with a center and scale shared over the whole group, a change in the score of a single response affects the entire coefficient vector. Moreover, the corresponding residuals appear as a within-group correlated holistic shift rather than mutually independent response-wise deviations. The uncertainty involved in training should thus be modeled for the entire coefficient vector, rather than for each response separately.
% 由 Eq.~\ref{eq:nominal_coefficient} 和 Eq.~\ref{eq:grpo_objective} 可知，组归一化使用共享的中心和尺度。因此，如 Fig.~\ref{fig:intro_joint_diagnostics}(a) 所示，由于 $T_{\mathrm{train}}$ 基于整组奖励共享的中心和尺度进行归一化，单个回答的评分变化会同时影响整个系数向量。此外，相应残差表现为组内相关的整体偏移，而非彼此独立的逐回答偏差，因此，训练所涉及的不确定性应针对整个系数向量建模，而不是分别处理各回答。

\textbf{Can Coefficient Drift Alone Characterize Update Safety?}
Fig.~\ref{fig:intro_joint_diagnostics}(b) applies the coefficients from the two scorings to the response-level gradients of the same response group and compares the local updates under different policy snapshots. The results show that similar coefficient shifts may produce different update magnitudes and directions, and that the updates induced by the two scorings may even have a negative inner product. Distance in coefficient space therefore cannot characterize update safety by itself, since the direction and magnitude of the update also depend on the gradient geometry of the current policy.

% Fig.~\ref{fig:intro_joint_diagnostics}(b) 将两次评分对应的系数作用于相同回答组的 response-level gradients，并在不同策略快照下比较局部更新。结果显示，相近的系数偏移可能产生不同的更新幅度和方向，部分第一次评分诱导的更新与第二次评分诱导的更新甚至具有负内积。因此，系数空间中的距离不能单独刻画更新安全性，更新的方向和大小还取决于当前策略的 gradient geometry。

\textbf{Can Group-wise Screening Control Aggregate Reversal Risk?}
The analyses in Fig.~\ref{fig:intro_joint_diagnostics} jointly indicate that when the optimizer retains only the response groups satisfying a local criterion, the sample distribution entering optimization changes from all response groups to the acceptance set. Since different screening rules yield different acceptance sets, a per-group criterion only shows that a local condition holds and cannot directly provide control over the aggregate direction-reversal risk on the acceptance set. Risk assessment must therefore be based on conditional statistics over the response groups that are actually accepted.

% Fig.~\ref{fig:intro_joint_diagnostics}的分析结果共同表明，当训练器只保留通过局部判据的回答组时，实际进入优化的样本分布由全部回答组变为接受集合。由于不同筛选规则会形成不同的接受集合，单组判据只能说明局部条件成立，不能直接给出接受集合上的总体方向反转风险控制。因此，风险评估必须针对实际被接受的回答组进行条件统计。

In summary, group normalization turns score discrepancies into correlated shifts in training coefficients, whose update effects depend on the current policy's response-level gradient geometry. Selective execution further changes the accepted groups and their aggregate direction-reversal risk. SetGRPO therefore jointly models coefficient uncertainty, policy-dependent update geometry, and acceptance-conditioned reversal risk.

% In summary, score discrepancies form within-group correlated coefficient shifts after group normalization, and the local updates induced by these shifts depend on the response-level gradient geometry of the current policy. When the optimizer further applies selective execution according to a local criterion, the acceptance set that actually enters training also changes, thereby affecting the aggregate direction-reversal risk. Policy optimization driven by scores must therefore handle jointly the joint uncertainty in coefficient space, the policy dependence in the update space, and the direction-reversal risk under acceptance. This motivates the design of SetGRPO developed below.
% 综上所述，评分差异经过组归一化后会形成组内相关的系数偏移，而这些偏移所诱导的局部更新又取决于当前策略的 response-level gradient geometry。当训练器进一步根据局部判据选择性执行回答组时，实际进入训练的接受集合也会随之变化，进而影响总体方向反转风险。因此，评分驱动的策略优化需要同时处理系数空间中的联合不确定性、更新空间中的策略依赖性和接受条件下的方向反转风险，这构成后文 SetGRPO 设计的出发点。

%% file: sec/4_method.tex
\section{Methodology}
\label{sec:setgrpo}

As the main innovation, SetGRPO constructs a joint coefficient set in the training-coefficient space actually used by GRPO and propagates it into the update space of the current policy, so as to recover effective and directionally consistent training updates as much as possible under grader score discrepancies. To address the propagation chain revealed by the pilot experiments in Section~\ref{sec:preliminaries}, SetGRPO consists of three interconnected modules: (a) \textit{Joint Reward Set Calibration, JRSC}, (b) \textit{Geometry-Guided Safe Recovery, GGSR}, and (c) \textit{Selective Risk-Gated Optimization, SRGO}. In the following, we introduce these three modules in turn.

% 作为本文的主要创新，SetGRPO 在 GRPO 实际使用的训练系数空间中构造联合系数集合，并将其传递到当前策略的更新空间，以在 grader 评分差异下尽可能恢复有效且方向一致的训练更新。针对第~\ref{sec:preliminaries}~节预实验揭示的传播链条，SetGRPO 由三个相互衔接的模块组成：(a) 联合奖励集合校准（\textit{Joint Reward Set Calibration, JRSC}）,(b) 几何引导的安全恢复（\textit{Geometry-Guided Safe Recovery, GGSR}）, and (c) 选择性风险门控优化（\textit{Selective Risk-Gated Optimization, SRGO}）。下面我们依次介绍这三个模块。

\subsection{Joint Reward Set Calibration (JRSC)}
\label{subsec:jrsc}
As shown in Fig.~\ref{fig:intro_joint_diagnostics}(a), two grader scorings of the same response group produce correlated within-group shifts after the shared training mapping. We therefore propose \textit{Joint Reward Set Calibration}, which models this coupled uncertainty as a joint coefficient set rather than independent response-wise errors. Specifically, for problem $x$, let $\mathbf r_x$ and $\mathbf r_x^\star$ denote the rewards from the first and second scorings. Using Eq.~\ref{eq:nominal_coefficient},
$\hat{\mathbf a}_x=T_{\mathrm{train}}(\mathbf r_x),\quad
\mathbf a_x^\star=T_{\mathrm{train}}(\mathbf r_x^\star),$
where $T_{\mathrm{train}}$ maps group rewards to within-group normalized coefficients. We define the paired residual
$\boldsymbol\delta_x=\mathbf a_x^\star-\hat{\mathbf a}_x$,
which calibrates the coverage radius of the subsequent joint set.

Subsequently, to capture how reward changes jointly perturb the coefficient vector, we linearize $T_{\mathrm{train}}$ around the first-scoring rewards $\mathbf r_x$ and define
$\mathbf J_x=\left.\nabla_{\mathbf r}T_{\mathrm{train}}(\mathbf r)\right|_{\mathbf r=\mathbf r_x}$.
Its $j$-th column describes how a perturbation in reward $j$ affects all group coefficients, preserving the cross-response coupling induced by shared normalization. The resulting coefficient-uncertainty matrix under unit reward perturbations is
% Subsequently, to describe the coupled effect of reward changes on the entire group of coefficients, we first linearize $T_{\mathrm{train}}$ to first order around the rewards $\mathbf r_x$ obtained from the first scoring, and define the Jacobian
% $\mathbf J_x=\left.\nabla_{\mathbf r}T_{\mathrm{train}}(\mathbf r)\right|_{\mathbf r=\mathbf r_x},$
% where the $j$-th column represents the effect of a local change in the $j$-th reward on the entire coefficient vector, preserving the cross-response coupling induced by within-group normalization. Based on the Jacobian, the coefficient uncertainty matrix under unit reward perturbations is
\begin{equation}
    \mathbf S_x^{\mathrm{coef}}=\mathbf J_x\mathbf J_x^\top+\lambda\mathbf I_K,
    \qquad \lambda\geq0,
\label{eq:jrsc_shape}
\end{equation}
% where $\mathbf J_x\mathbf J_x^\top$ aggregates the coupled effect of reward perturbations on the whole group of coefficients under the local linearization and determines the allowable scale of shift in different directions of the joint coefficient set, while $\lambda\mathbf I_K$ is a numerical regularization term for the case of matrix degeneracy, used to stabilize matrix inversion and distance computations.
where $\mathbf J_x\mathbf J_x^\top$ captures the cross-coefficient effects of local reward perturbations and determines set's directional scales, while $\lambda\mathbf I_K$ stabilizes inversion and distance computation in near-degenerate cases.

% 随后，为描述 reward 变化对整组系数的联动影响，我们首先在当前第一次评分得到的 reward $\mathbf r_x$ 处将 $T_{\mathrm{train}}$ 一阶线性化，并定义 Jacobian
% $\mathbf J_x=\left.\nabla_{\mathbf r}T_{\mathrm{train}}(\mathbf r)\right|_{\mathbf r=\mathbf r_x},$
% 其中，第 $j$ 列表示第 $j$ 个 reward 的局部变化对整个系数向量的影响，以保留组归一化产生的跨回答耦合。基于Jacobian，我们得到单位 reward 扰动下的系数不确定性矩阵为
% \begin{equation}
%     \mathbf S_x^{\mathrm{coef}}=\mathbf J_x\mathbf J_x^\top+\lambda\mathbf I_K,
%     \qquad \lambda\geq0,
% \label{eq:jrsc_shape}
% \end{equation}
% 其中，$\mathbf J_x\mathbf J_x^\top$ 汇总了局部线性化下各 reward 扰动对整组系数的联动影响，并决定联合系数集合在不同方向上允许的偏移尺度，$\lambda\mathbf I_K$ 为矩阵退化时的数值正则项，用于稳定求逆和距离计算。

However, $\mathbf S_z^{\mathrm{coef}}$ specifies only the shape of the joint coefficient set, not its radius. We therefore calibrate a unified coverage radius from independent paired residuals. For each coverage-calibration group $z\in\mathcal D_{\mathrm{cov}}$
\footnote{The paired groups are split by problem root into disjoint coverage- and risk-calibration sets. The former calibrates the set radius, while the latter selects the execution threshold; both are fixed before formal training.}
, we compute the Mahalanobis score~\citep{mahalanobis1936generalised}
$\rho_z=\sqrt{\boldsymbol\delta_z^\top
(\mathbf S_z^{\mathrm{coef}})^{-1}\boldsymbol\delta_z}.$
This score accounts for directional scale differences and within-group correlations induced by reward normalization.

Next, let $n=|\mathcal D_{\mathrm{cov}}|$, $\rho_{(j)}$ denote the $j$-th order statistic, and $\alpha\in(0,1)$ be the target miss rate. Under exchangeability of the calibration and target groups  (i.e., identical data construction, response sampling, and scoring protocols with permutation-invariant joint distributions), we define
% Next, let $n=|\mathcal D_{\mathrm{cov}}|$, $\rho_{(j)}$ denote the $j$-th order statistic, and $\alpha\in(0,1)$ be the target miss rate. To ensure that the quantile admits a finite-sample coverage guarantee, we assume that the coverage calibration groups and the group to be covered are exchangeable, that is, under the same data construction, response sampling, and scoring protocol, their joint distribution is invariant to any permutation. Under this condition, set
\begin{equation}
% \begin{aligned}
\small
    Q_\alpha=\rho_{\left(\left\lceil(n+1)(1-\alpha)\right\rceil\right)},\quad
    \mathcal C_x=\left\{\mathbf a\in\mathbb R^K:(\mathbf a-\hat{\mathbf a}_x)^\top
    (\mathbf S_x^{\mathrm{coef}})^{-1}(\mathbf a-\hat{\mathbf a}_x)\leq Q_\alpha^2\right\}.
% \end{aligned}
\label{eq:jrsc_set}
\end{equation}
Then $\mathcal C_x$ is an ellipsoidal joint coefficient set centered at $\hat{\mathbf a}_x$ that contains the training coefficients of the second scoring with marginal probability at least $1-\alpha$.
\textcolor{black}{If $\left\lceil(n+1)(1-\alpha)\right\rceil=n+1$, no corresponding $(n+1)$-th order statistic exists in the calibration set. }
% 随后，记 $n=|\mathcal D_{\mathrm{cov}}|$，$\rho_{(j)}$ 为第 $j$ 个次序统计量，$\alpha\in(0,1)$ 为目标误覆盖率。为了使分位数具有有限样本覆盖保证，我们假设覆盖校准组与待覆盖回答组满足 exchangeability，即在相同的数据构造、回答采样和评分协议下，这些回答组的联合分布对任意置换保持不变。在该条件下，令
% \begin{equation}
% % \begin{aligned}
% \small
%     Q_\alpha=\rho_{\left(\left\lceil(n+1)(1-\alpha)\right\rceil\right)},\quad
%     \mathcal C_x=\left\{\mathbf a\in\mathbb R^K:(\mathbf a-\hat{\mathbf a}_x)^\top
%     (\mathbf S_x^{\mathrm{coef}})^{-1}(\mathbf a-\hat{\mathbf a}_x)\leq Q_\alpha^2\right\}.
% % \end{aligned}
% \label{eq:jrsc_set}
% \end{equation}
% 则 $\mathcal C_x$ 是以 $\hat{\mathbf a}_x$ 为中心的椭球联合系数集合，并以至少 $1-\alpha$ 的边际概率包含第二次评分对应的训练系数。
% \textcolor{black}{若 $\left\lceil(n+1)(1-\alpha)\right\rceil=n+1$，则校准集不存在相应的第 $n+1$ 个次序统计量；此时按附录~\ref{app:finite_sample_coverage} 的保守约定取 $Q_\alpha=+\infty$。}
% \textcolor{black}{该椭球集合的有限样本覆盖证明见附录~\ref{app:finite_sample_coverage}。}

\begin{figure*}[t]
    \centering
            \vspace{-12pt}
    \includegraphics[
        width=\textwidth,
        keepaspectratio
    ]{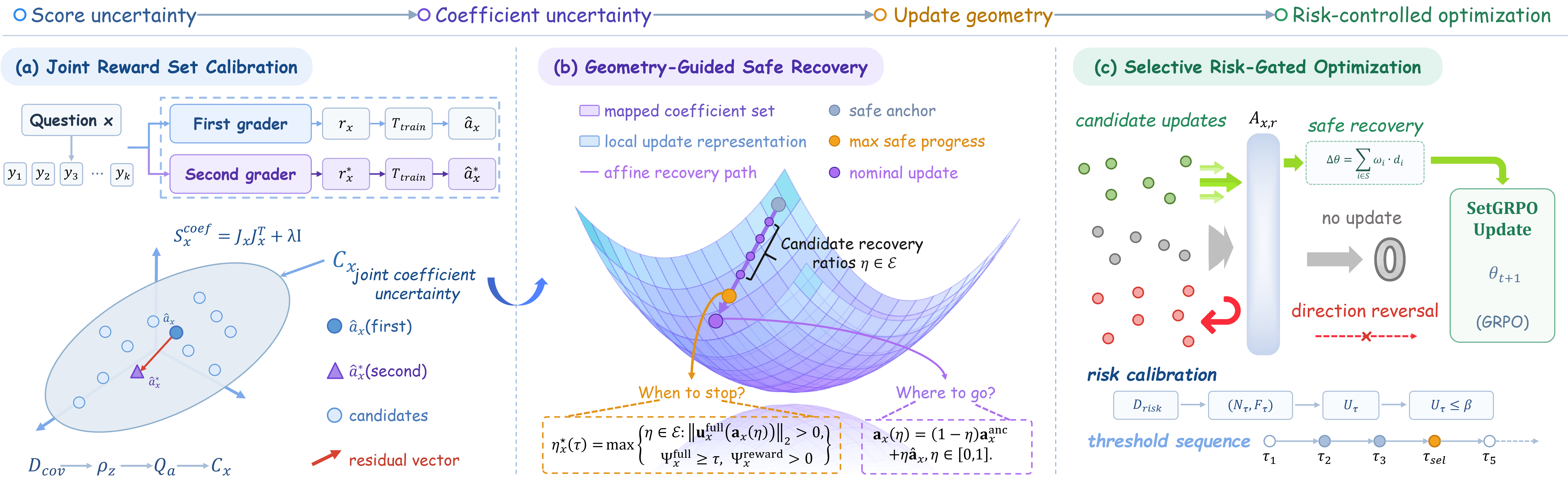}
    \vspace{-22pt}
    \caption{Overview of SetGRPO. 
    \textit{(a) Joint reward-set calibration} uses two grader scores on the same response group to construct a joint coefficient uncertainty set. \textit{(b) Geometry-guided safe recovery} maps candidate coefficients through the current policy geometry and identifies the largest direction-preserving recovery ratio. \textit{(c) Selective risk-gated optimization} selects a global threshold and applies the recovered coefficients only to updates satisfying the calibrated conditional-risk constraint.}
    \label{fig:setgrpo_framework}
        \vspace{-18pt}
\end{figure*}

\subsection{Geometry-Guided Safe Recovery (GGSR)}
\label{subsec:ggsp}

JRSC models the discrepancy between two grader scorings as the joint coefficient set $\mathcal C_x$. We then propose \textit{Geometry-Guided Safe Recovery (GGSR)}, which maps $\mathcal C_x$ into the policy-update and reward spaces to identify the largest coefficient-recovery ratio whose update remains directionally consistent with the second scoring under the current policy.

% JRSC represents the change in training coefficients caused by two grader scorings as the joint set $\mathcal C_x$. To decide which candidate coefficients may be used for training, however, it remains necessary to examine the updates that they actually induce under the current policy. To this end, we design \textit{Geometry-Guided Safe Recovery}, which maps $\mathcal C_x$ into the policy-update space and the reward space and checks whether a candidate update can remain directionally consistent with the update induced by the second scoring under score discrepancies, thereby determining the largest proportion of coefficients that can be safely recovered for each response group.
% JRSC 将两次 grader 评分造成的训练系数变化表示为联合集合 $\mathcal C_x$，但要决定哪些候选系数可以用于训练，还需要考察它们在当前策略下诱导的实际更新。为此，我们设计了几何引导的安全恢复（Geometry-Guided Safe Recovery, GGSR）：它将 $\mathcal C_x$ 映射到策略更新空间和 reward space，并检验候选更新在评分差异下能否与第二次评分对应的更新保持方向一致，从而确定每个回答组可安全恢复的最大系数比例。

% Specifically, we first let $q\in\{\mathrm{full},\mathrm{reward}\}$ denote the policy-update space and the reward space, respectively, with dimension $D_q$; let $\mathcal S^q$ be a fixed low-dimensional linear map, $\mathbf P_x^q$ a linear transformation, and $\mathbf h_x^q$ an offset independent of the coefficients. Notably, by full geometry we mean the complete local update geometry under a fixed projected representation, rather than the unprojected original parameter space. From the response-level gradients in Eq.~\ref{eq:local_update_preliminary}, we define
Specifically, we first let $q\in\{\mathrm{full},\mathrm{reward}\}$ denote the policy-update and reward spaces, respectively, with dimension $D_q$. Let $\mathcal S^q$ be a fixed low-dimensional linear map, $\mathbf P_x^q$ a linear transformation, and $\mathbf h_x^q$ a coefficient-independent offset. Notably, by full geometry we mean the complete local update geometry under a fixed projected representation, rather than the unprojected original parameter space. From the response-level gradients in Eq.~\ref{eq:local_update_preliminary}, we define
\begin{equation}
% \begin{aligned}
\small
    \mathbf G_x^q=\left[\mathcal S^q(\mathbf g_{x,1}),\ldots,
    \mathcal S^q(\mathbf g_{x,K})\right]\in\mathbb R^{D_q\times K},\quad
    \mathbf u_x^q(\mathbf a)=\mathbf P_x^q\left(\mathbf G_x^q\mathbf a+\mathbf h_x^q\right)\in\mathbb R^{D_q},
% \end{aligned}
\label{eq:ggsp_geometry}
\end{equation}
where the $i$-th column of $\mathbf G_x^q$ is the projected gradient of response $y_{x,i}$, and $\mathbf u_x^q(\mathbf a)$ is the local update induced by the coefficients $\mathbf a$ in the space $q$.

% 具体而言，我们首先 记 $q\in\{\mathrm{full},\mathrm{reward}\}$ 分别表示策略更新空间和 reward space，维数为 $D_q$；$\mathcal S^q$ 为固定的低维线性映射，$\mathbf P_x^q$ 为线性变换，$\mathbf h_x^q$ 为与系数无关的偏移项。值得注意的是，我们所称的 full geometry 是固定投影表示下的完整局部更新几何，而不是未经投影的原始参数空间。由 Eq.~\ref{eq:local_update_preliminary} 中的回答级梯度，我们可定义
% \begin{equation}
% % \begin{aligned}
%     \mathbf G_x^q=\left[\mathcal S^q(\mathbf g_{x,1}),\ldots,
%     \mathcal S^q(\mathbf g_{x,K})\right]\in\mathbb R^{D_q\times K},\quad
%     \mathbf u_x^q(\mathbf a)=\mathbf P_x^q\left(\mathbf G_x^q\mathbf a+\mathbf h_x^q\right)\in\mathbb R^{D_q},
% % \end{aligned}
% \label{eq:ggsp_geometry}
% \end{equation}
% 其中，$\mathbf G_x^q$ 的第 $i$ 列是回答 $y_{x,i}$ 的投影梯度，$\mathbf u_x^q(\mathbf a)$ 是系数 $\mathbf a$ 在空间 $q$ 中诱导的局部更新。

For a candidate coefficient $\mathbf a$, let $\mathbf d=\mathbf u_x^q(\mathbf a)$ and define the worst-case projection over the joint coefficient set as
$\Psi_x^q(\mathbf d)
:=\min_{\tilde{\mathbf a}\in\mathcal C_x}
\mathbf d^\top\mathbf u_x^q(\tilde{\mathbf a}).$
This quantity measures the candidate update's worst-case alignment and magnitude under grader uncertainty. If $\mathbf a_x^\star\in\mathcal C_x$, then
$\Psi_x^{\mathrm{full}}\!\left(\mathbf u_x^{\mathrm{full}}(\mathbf a)\right)
\le
\left\langle
\mathbf u_x^{\mathrm{full}}(\mathbf a),
\mathbf u_x^{\mathrm{full}}(\mathbf a_x^\star)
\right\rangle.$
Thus, a positive full-space projection rules out direction reversal, while a positive reward-space projection preserves reward-direction consistency. Because $\mathcal C_x$ is ellipsoidal and $\mathbf u_x^q$ is affine in $\mathbf a$, the projection has the closed form
\begin{equation}
\begin{aligned}
    \Psi_x^q(\mathbf d)
    =\mathbf d^\top\mathbf u_x^q(\hat{\mathbf a}_x)
    -Q_\alpha\sqrt{
    \mathbf d^\top\mathbf P_x^q\mathbf G_x^q\mathbf S_x^{\mathrm{coef}}
    (\mathbf G_x^q)^\top(\mathbf P_x^q)^\top\mathbf d}.
\end{aligned}
\label{eq:ggsp_margin}
\end{equation}

However, the minimum-projection criterion evaluates a candidate coefficient but does not select one from $\mathcal C_x$. We therefore choose the feasible coefficient with the smallest local-update norm as an anchor, denoted as
$\mathbf a_x^{\mathrm{anc}}\in\arg\min_{\mathbf a\in\mathcal C_x}
\frac{1}{2}\|\mathbf u_x^{\mathrm{full}}(\mathbf a)\|_2^2.$
Since $\mathcal C_x$ is convex and centered at $\hat{\mathbf a}_x$, the path
$\mathbf a_x(\eta)=(1-\eta)\mathbf a_x^{\mathrm{anc}}+\eta\hat{\mathbf a}_x,
\qquad \eta\in[0,1],$
remains in $\mathcal C_x$. Here, $\eta$ denotes the fraction of training recovered from the conservative anchor toward the first-scoring update. For a predetermined candidate set $\mathcal E\subseteq[0,1]$, the maximum acceptable ratio under threshold $\tau\geq0$ is
% However, this minimum projection criterion only tests a given candidate coefficient and does not directly determine which coefficient the optimizer should adopt from $\mathcal C_x$. To this end, we first select the feasible coefficient with the smallest local update norm as an anchor, denoted
% $\mathbf a_x^{\mathrm{anc}}\in\arg\min_{\mathbf a\in\mathcal C_x}
% \frac{1}{2}\|\mathbf u_x^{\mathrm{full}}(\mathbf a)\|_2^2.$
% Since $\mathcal C_x$ is a convex set and $\hat{\mathbf a}_x$ is its center, the interpolation from this anchor to $\hat{\mathbf a}_x$,
% $\mathbf a_x(\eta)=(1-\eta)\mathbf a_x^{\mathrm{anc}}+\eta\hat{\mathbf a}_x,
% \quad \eta\in[0,1],$
% lies inside $\mathcal C_x$ for all $\eta$; $\eta$ is the proportion of training recovered from the conservative update toward the first-scoring update. Let $\mathcal E\subseteq[0,1]$ be a predetermined set of candidate ratios. Then the maximum acceptable ratio under a threshold $\tau\geq0$ is
\begin{equation}
\fontsize{8pt}{8pt}
    \eta_x^\star(\tau)=\max\left\{\eta\in\mathcal E:
    \left\|\mathbf u_x^{\mathrm{full}}(\mathbf a_x(\eta))\right\|_2>0,
    \;\Psi_x^{\mathrm{full}}\!\left(\mathbf u_x^{\mathrm{full}}(\mathbf a_x(\eta))\right)\geq\tau,
    \;\Psi_x^{\mathrm{reward}}\!\left(\mathbf u_x^{\mathrm{reward}}(\mathbf a_x(\eta))\right)>0
    \right\}.
\label{eq:ggsp_recovery}
\end{equation}
If no $\eta$ satisfies these conditions, group $x$ produces no update.

% If no $\eta$ satisfies these conditions, the response group $x$ yields no recovery update. \textcolor{black}{A proof of the existence of the anchor and of the fact that the linear recovery path above always lies inside $\mathcal C_x$ is given in Appendix~\ref{app:recovery_path_proof}.}

% 然而，该最小投影判据只能检验给定候选系数，不能直接决定训练器应从 $\mathcal C_x$ 中采用哪个系数。为此，我们先选取局部更新范数最小的可行系数作为锚点，记作
% $\mathbf a_x^{\mathrm{anc}}\in\arg\min_{\mathbf a\in\mathcal C_x}
% \frac{1}{2}\|\mathbf u_x^{\mathrm{full}}(\mathbf a)\|_2^2.$
% 由于 $\mathcal C_x$ 是凸集且 $\hat{\mathbf a}_x$ 为其中心，从该锚点到 $\hat{\mathbf a}_x$ 的插值
% $\mathbf a_x(\eta)=(1-\eta)\mathbf a_x^{\mathrm{anc}}+\eta\hat{\mathbf a}_x,
% \quad \eta\in[0,1],$
% 始终位于 $\mathcal C_x$ 内；$\eta$ 表示从保守更新向第一次评分更新恢复的训练比例。令 $\mathcal E\subseteq[0,1]$ 为预先固定的候选比例集合，则阈值 $\tau\geq0$ 下的最大可接受比例为
% \begin{equation}
% \small
%     \eta_x^\star(\tau)=\max\left\{\eta\in\mathcal E:
%     \left\|\mathbf u_x^{\mathrm{full}}(\mathbf a_x(\eta))\right\|_2>0,
%     \;\Psi_x^{\mathrm{full}}\!\left(\mathbf u_x^{\mathrm{full}}(\mathbf a_x(\eta))\right)\geq\tau,
%     \;\Psi_x^{\mathrm{reward}}\!\left(\mathbf u_x^{\mathrm{reward}}(\mathbf a_x(\eta))\right)>0
%     \right\}.
% \label{eq:ggsp_recovery}
% \end{equation}
% 若不存在满足条件的 $\eta$，回答组 $x$ 不产生恢复更新。\textcolor{black}{锚点的存在性以及上述线性恢复路径始终位于 $\mathcal C_x$ 内的证明见附录~\ref{app:recovery_path_proof}。}

\subsection{Selective Risk-Gated Optimization (SRGO)}
\label{subsec:srgo}

Equation~\ref{eq:ggsp_recovery} determines the maximum feasible recovery ratio for each response group at threshold $\tau$. Because $\tau$ also controls whether a nonzero recovery exists, it induces the acceptance set
$\mathcal A_\tau=\{x:A_{x,\tau}=1\},$
where $A_{x,\tau}=1$ iff the constraints admit a nonzero recovery ratio, and $A_{x,\tau}=0$ otherwise. The aggregate direction-reversal risk therefore depends on the composition of $\mathcal A_\tau$, not only on the group-wise criterion. We thus introduce \textit{Selective Risk-Gated Optimization}, which evaluates acceptance-conditioned risk for candidate thresholds on the independent risk-calibration set $\mathcal D_{\mathrm{risk}}$ and selects a threshold that satisfies the risk constraint while retaining as much training signal as possible.

Specifically, for each candidate threshold $\tau\in\mathcal T$, we define the direction-reversal event for accepted groups as
$Z_{x,\tau}
=\mathbb I\!\left[
\left\langle
\mathbf u_x^{\mathrm{full}}\!\left(\mathbf a_x(\eta_x^\star(\tau))\right),
\mathbf u_x^{\mathrm{full}}(\mathbf a_x^\star)
\right\rangle<0
\right],$
and its acceptance-conditioned risk as
$R_\tau=\Pr(Z_{x,\tau}=1\mid A_{x,\tau}=1).$
On the independent risk-calibration set $\mathcal D_{\mathrm{risk}}$, let $N_\tau$ and $F_\tau$ denote the numbers of accepted and direction-reversed groups, respectively. We estimate $R_\tau$ using the one-sided Clopper--Pearson upper bound
$U_\tau=\operatorname{CPUpper}
\left(F_\tau,N_\tau;1-\frac{\gamma}{|\mathcal T|}\right),$
where $\beta$ is the target risk level and $\gamma$ is the family-wise failure probability across candidate thresholds. A threshold is feasible if $N_\tau>0$ and $U_\tau\leq\beta$; among feasible thresholds, we select the one maximizing cumulative recovery $Y_\tau$, formualted as,
\begin{equation}
\small
    Y_\tau=\sum_{x\in\mathcal D_{\mathrm{risk}}}A_{x,\tau}\eta_x^\star(\tau),\quad
    \tau_{\mathrm{sel}}=\min\left(
    \arg\max_{\substack{\tau\in\mathcal T:N_\tau>0,\;U_\tau\leq\beta}}Y_\tau\right).
\label{eq:srgo_selection}
\end{equation}
If several thresholds tie, the smaller one is taken; if no threshold satisfies the risk constraint, $\tau_{\mathrm{sel}}$ is undefined and the risk gating rule is disabled in subsequent training.

% 若多个阈值并列，则取较小者；若不存在满足风险约束的阈值，则不定义 $\tau_{\mathrm{sel}}$，后续训练不启用该风险门控规则。

Once $\tau_{\mathrm{sel}}$ is determined, SetGRPO retains the probability ratio, clipping, and token-level surrogate term of the original GRPO and replaces only the training coefficients with,
% 确定 $\tau_{\mathrm{sel}}$ 后，SetGRPO 保留原始 GRPO 的 probability ratio、clipping 和 token-level surrogate term，仅将训练系数替换为
\begin{equation}
\small
    \mathcal J_{\mathrm{SetGRPO}}(\boldsymbol\theta)=
    \mathcal J_{\mathrm{GRPO}}\!\left(\boldsymbol\theta;\widetilde{\mathbf a}_x\right),
    \qquad
    \widetilde{\mathbf a}_x=
    \begin{cases}
        \mathbf a_x\!\left(\eta_x^\star(\tau_{\mathrm{sel}})\right), & A_{x,\tau_{\mathrm{sel}}}=1,\\
        \mathbf 0_K, & A_{x,\tau_{\mathrm{sel}}}=0.
    \end{cases}
\label{eq:setgrpo_final}
\end{equation}
Accepted response groups use their recovered coefficients, while the remaining groups yield no policy update. We further evaluate the predetermined $\tau_{\mathrm{sel}}$ across different training checkpoints to test its empirical stability under policy drift.
% 接受的回答组使用恢复系数，其余回答组不产生策略更新。我们进一步在不同训练 checkpoint 上评估预先确定的 $\tau_{\mathrm{sel}}$，以检验策略变化下该阈值的经验稳定性，详细结果见附录
% ~\ref{app:policy_drift_validation}。

%% file: sec/5_experiment.tex
% !TEX root = ../neurips_2025.tex

% =========================
% Table colors and commands
% =========================
\definecolor{setgrpoBlue}{HTML}{EEF2FF}

% Green intensity indicates the magnitude of improvement.
\definecolor{gainLow}{HTML}{69A875}   % gain < 4.0
\definecolor{gainMid}{HTML}{258A50}   % 4.0 <= gain < 7.0
\definecolor{gainHigh}{HTML}{006B3C}  % gain >= 7.0

% Only the improvement value is colored; the score and parentheses remain black.
\newcommand{\tabgainL}[2]{%
  #1{\scriptsize\,(\textcolor{gainLow}{#2})}%
}
\newcommand{\tabgainM}[2]{%
  #1{\scriptsize\,(\textcolor{gainMid}{#2})}%
}
\newcommand{\tabgainH}[2]{%
  #1{\scriptsize\,(\textcolor{gainHigh}{#2})}%
}

% =========================
% Main results table
% =========================
\begin{table*}[t]
\centering
\vspace{-3pt}
\caption{
Main results on ten downstream benchmarks.
HB: HealthBench; LEM: LLMEval-Med; MQA: MedQA;
MCQA: MedMCQA; WB: WritingBench; CW: Creative Writing v3;
IFE: IFEval; IFB: IFBench; IB: InfoBench; and MIF: MulDimIF.
Our results report the mean across repeated runs, with absolute improvements over the corresponding Base model shown in parentheses.
}
\vspace{-10pt}
\label{tab:main_results}

\renewcommand{\arraystretch}{1.08}
\setlength{\tabcolsep}{2pt}
\scriptsize

\resizebox{\textwidth}{!}{%
\begin{tabular}{@{}lccccccccccc@{}}
\toprule

\multirow{2}{*}{\textbf{Model}}
& \multicolumn{4}{c}{\textbf{Medical}}
& \multicolumn{2}{c}{\textbf{Writing}}
& \multicolumn{4}{c}{\textbf{Instruction Following}}
& \multirow{2}{*}{\textbf{Macro Avg.}}
\\
\cmidrule(lr){2-5}
\cmidrule(lr){6-7}
\cmidrule(lr){8-11}

& \textbf{HB}
& \textbf{LEM}
& \textbf{MQA}
& \textbf{MCQA}
& \textbf{WB}
& \textbf{CW}
& \textbf{IFE}
& \textbf{IFB}
& \textbf{IB}
& \textbf{MIF}
&
\\
\midrule

\multicolumn{12}{c}{\textbf{Closed-source Models}}
\\
\midrule

OpenAI-o3
& 59.80
& 78.90
& 96.00
& 84.70
& 77.70
& 81.40
& 91.60
& 67.80
& 89.40
& 64.00
& 79.13
\\

GPT-4.1
& 47.90
& 71.20
& 92.40
& 80.00
& 69.00
& 79.00
& 87.00
& 37.20
& 89.40
& 64.00
& 71.71
\\

Gemini-3 Pro Preview
& 49.30
& 72.70
& 67.00
& 62.20
& 78.50
& 81.50
& 94.20
& 61.20
& 85.60
& 66.83
& 71.90
\\

\midrule
\multicolumn{12}{c}{\textbf{Open-sourced Models}}
\\
\midrule

Qwen3.5-4B
& 37.09
& 26.24
& 77.77
& 63.26
& 50.20
& 27.60
& 46.40
& 17.35
& 78.22
& 41.25
& 46.54
\\

\quad + GRPO
& \tabgainL{56.94}{+19.85}
& \tabgainL{43.29}{+17.05}
& \tabgainL{79.11}{+1.34}
& \tabgainL{63.55}{+0.29}
& \tabgainL{64.92}{+14.72}
& \tabgainL{29.98}{+2.38}
& \tabgainM{62.28}{+15.88}
& \tabgainM{25.09}{+7.74}
& \tabgainM{80.38}{+2.16}
& \tabgainM{48.55}{+7.30}
& \tabgainL{55.41}{+8.87}
\\

\quad + RuscaRL
& \tabgainM{65.10}{+28.01}
& \tabgainM{45.30}{+19.06}
& \tabgainM{79.60}{+1.83}
& \tabgainM{63.30}{+0.04}
& \tabgainM{67.50}{+17.30}
& \tabgainM{31.80}{+4.20}
& \tabgainM{65.90}{+19.50}
& \tabgainM{24.40}{+7.05}
& \tabgainM{80.74}{+2.52}
& \tabgainM{47.80}{+6.55}
& \tabgainM{57.14}{+10.60}
\\

\quad + RubricHub
& \tabgainL{67.18}{+30.09}
& \tabgainM{47.71}{+21.47}
& \tabgainL{78.06}{+0.29}
& \textbf{\tabgainL{64.22}{+0.96}}
& \tabgainM{69.62}{+19.42}
& \textbf{\tabgainM{32.80}{+5.20}}
& \tabgainM{67.88}{+21.48}
& \tabgainM{26.31}{+8.96}
& \textbf{\tabgainM{82.91}{+4.69}}
& \tabgainM{49.84}{+8.59}
& \tabgainM{58.65}{+12.11}
\\

\rowcolor{setgrpoBlue}
\textbf{\quad + SetGRPO (Ours)}
& \textbf{\tabgainH{69.64}{+32.55}}
& \textbf{\tabgainH{50.08}{+23.84}}
& \textbf{\tabgainH{80.55}{+2.78}}
& \tabgainH{63.84}{+0.58}
& \textbf{\tabgainH{72.14}{+21.94}}
& \tabgainH{32.23}{+4.63}
& \textbf{\tabgainH{70.45}{+24.05}}
& \textbf{\tabgainH{28.97}{+11.62}}
& \tabgainH{82.32}{+4.10}
& \textbf{\tabgainH{52.35}{+11.10}}
& \textbf{\tabgainH{60.26}{+13.72}}
\\

\midrule

Qwen3.5-9B
& 40.94
& 34.63
& 76.20
& 67.03
& 53.83
& 32.19
& 51.39
& 19.39
& 80.98
& 41.17
& 49.78
\\

\quad + GRPO
& \tabgainL{61.06}{+20.12}
& \tabgainL{51.65}{+17.02}
& \tabgainL{77.68}{+1.48}
& \tabgainL{67.10}{+0.07}
& \tabgainL{68.84}{+15.01}
& \tabgainL{34.65}{+2.46}
& \tabgainM{67.61}{+16.22}
& \tabgainM{26.91}{+7.52}
& \tabgainM{83.02}{+2.04}
& \tabgainM{48.33}{+7.16}
& \tabgainL{58.69}{+8.91}
\\

\quad + RuscaRL
& \tabgainM{69.00}{+28.06}
& \tabgainM{53.50}{+18.87}
& \tabgainM{78.35}{+2.15}
& \tabgainM{67.18}{+0.15}
& \tabgainM{71.70}{+17.87}
& \tabgainM{36.40}{+4.21}
& \tabgainM{71.30}{+19.91}
& \tabgainM{25.80}{+6.41}
& \tabgainM{82.86}{+1.88}
& \tabgainM{47.20}{+6.03}
& \tabgainM{60.33}{+10.55}
\\

\quad + RubricHub
& \tabgainL{71.04}{+30.10}
& \tabgainM{55.62}{+20.99}
& \textbf{\tabgainL{79.86}{+3.66}}
& \tabgainL{67.28}{+0.25}
& \tabgainM{73.76}{+19.93}
& \tabgainM{34.21}{+2.02}
& \tabgainM{73.38}{+21.99}
& \tabgainM{28.24}{+8.85}
& \textbf{\tabgainM{85.42}{+4.44}}
& \textbf{\tabgainM{52.33}{+11.16}}
& \tabgainM{62.11}{+12.33}
\\

\rowcolor{setgrpoBlue}
\textbf{\quad + SetGRPO (Ours)}
& \textbf{\tabgainH{73.86}{+32.92}}
& \textbf{\tabgainH{58.15}{+23.52}}
& \tabgainH{79.32}{+3.12}
& \textbf{\tabgainH{67.34}{+0.31}}
& \textbf{\tabgainH{76.28}{+22.45}}
& \textbf{\tabgainH{36.91}{+4.72}}
& \textbf{\tabgainH{75.94}{+24.55}}
& \textbf{\tabgainH{30.57}{+11.18}}
& \tabgainH{84.83}{+3.85}
& \tabgainH{51.87}{+10.70}
& \textbf{\tabgainH{63.51}{+13.73}}
\\

\midrule

Llama 3.2-3B
& 14.37
& 15.96
& 57.18
& 48.74
& 30.63
& 32.57
& 71.08
& 23.67
& 44.93
& 35.69
& 37.48
\\

\quad + GRPO
& \tabgainL{21.08}{+6.71}
& \tabgainL{17.42}{+1.46}
& \tabgainL{58.66}{+1.48}
& \tabgainL{49.32}{+0.58}
& \tabgainL{40.49}{+9.86}
& \tabgainL{35.41}{+2.84}
& \tabgainL{77.69}{+6.61}
& \tabgainL{25.58}{+1.91}
& \tabgainL{45.14}{+0.21}
& \tabgainL{76.42}{+40.73}
& \tabgainL{44.72}{+7.24}
\\

\quad + RuscaRL
& \tabgainM{22.84}{+8.47}
& \tabgainM{17.96}{+2.00}
& \tabgainM{60.47}{+3.29}
& \tabgainM{51.05}{+2.31}
& \tabgainM{42.52}{+11.89}
& \tabgainM{40.28}{+7.71}
& \tabgainM{79.58}{+8.50}
& \tabgainM{27.76}{+4.09}
& \tabgainM{46.67}{+1.74}
& \tabgainM{79.35}{+43.66}
& \tabgainM{46.85}{+9.37}
\\

\quad + RubricHub
& \tabgainM{24.61}{+10.24}
& \tabgainM{18.19}{+2.23}
& \tabgainM{61.34}{+4.16}
& \tabgainM{51.27}{+2.53}
& \tabgainM{47.33}{+16.70}
& \tabgainM{39.84}{+7.27}
& \tabgainM{80.93}{+9.85}
& \tabgainM{28.67}{+5.00}
& \tabgainM{47.44}{+2.51}
& \tabgainM{82.08}{+46.39}
& \tabgainM{48.17}{+10.69}
\\

\rowcolor{setgrpoBlue}
\textbf{\quad + SetGRPO (Ours)}
& \textbf{\tabgainH{26.72}{+12.35}}
& \textbf{\tabgainH{18.47}{+2.51}}
& \textbf{\tabgainH{62.18}{+5.00}}
& \textbf{\tabgainH{52.14}{+3.40}}
& \textbf{\tabgainH{45.66}{+15.03}}
& \textbf{\tabgainH{42.51}{+9.94}}
& \textbf{\tabgainH{83.47}{+12.39}}
& \textbf{\tabgainH{30.38}{+6.71}}
& \textbf{\tabgainH{49.46}{+4.53}}
& \textbf{\tabgainH{84.67}{+48.98}}
& \textbf{\tabgainH{49.57}{+12.09}}
\\

\bottomrule
\end{tabular}%
}
\vspace{-10pt}
\end{table*}

\section{Experiments}
\label{sec:experiments}

\subsection{Experimental Setup}
\label{subsec:experimental_setup}

\textbf{Data and evaluation.}
We perform reinforcement learning training on the RubricHub training set~\citep{li2026rubrichub}, whose data covers five task types: Chat, Instruction Following, Medical, Science, and Writing. After training, each model is evaluated on ten downstream benchmarks spanning three domains: healthcare, writing, and instruction following. The healthcare benchmarks include HealthBench, LLMEval-Med, MedQA, and MedMCQA; the writing benchmarks include WritingBench and Creative Writing v3; and the instruction following benchmarks include IFEval, IFBench, InfoBench, and MulDimIF. For benchmarks that require model scoring, we uniformly invoke the GPT-5.6-Luna API. Each benchmark uses its official evaluation metrics, and Macro Avg. is used only to aggregate the results of the ten benchmarks.

% 我们使用 RubricHub 训练集~\citep{li2026rubrichub}进行强化学习训练，训练数据覆盖 Chat、Instruction Following、Medical、Science 和 Writing 五类任务。训练完成后，模型在十个下游 benchmark 上评估，这些 benchmark 涵盖医疗、写作和指令遵循三个领域：医疗领域包括 HealthBench、LLMEval-Med、MedQA 和 MedMCQA，写作领域包括 WritingBench 和 Creative Writing v3，指令遵循领域包括 IFEval、IFBench、InfoBench 和 MulDimIF。对于需要模型评分的下游 benchmark，我们统一调用 GPT-5.6-Luna API。各 benchmark 采用官方评测指标，Macro Avg. 仅用于汇总十个 benchmark 的结果。

% 数据按问题根划分为训练集、覆盖校准集和风险校准集。覆盖校准集用于确定联合系数集合的覆盖范围，风险校准集用于选择执行阈值；十个下游 benchmark 只在训练完成后用于最终评测，不参与训练、校准或 checkpoint 选择。不同 backbone 分别生成回答组并完成校准，不共享训练系数、Jacobian 或执行阈值。

\textbf{Models and baselines.}
We use Qwen3.5-4B, Qwen3.5-9B and Llama 3.2-3B as backbones and compare Base (the backbone without the training in this study) against four training methods: Rubric-based GRPO, RubricHub~\citep{li2026rubrichub}, RuscaRL~\citep{zhou2025ruscarl}, and SetGRPO. Rubric-based GRPO uses the same rewards obtained from the first scoring as SetGRPO and is the main control for differences in the training objective; RubricHub and RuscaRL are run following their publicly available training procedures, with identical data scale, training budget, and final evaluation protocol. 
% 我们以 Qwen3.5-4B 和 Qwen3.5-9B 为 backbone，比较 Base (表示未经过本实验训练的 backbone) 与四种训练方法：Rubric-based GRPO、RubricHub、RuscaRL 和 SetGRPO。Rubric-based GRPO 使用与 SetGRPO 相同的第一次评分得到的 reward，是控制训练目标差异的主要对照；RubricHub~\citep{li2026rubrichub} 和 RuscaRL\cite{zhou2025ruscarl} 按其公开训练流程运行，并使用相同的数据规模、训练预算和最终评测协议。具体训练设置见 Appendix~\ref{app:implementation_details}。

\textbf{Training details.}
For each prompt, we generate $K=8$ rollout responses with a maximum response length of 4096 tokens. All methods use the same training data, number of training steps, and generation budget, and are evaluated under the same protocol after training. For benchmarks requiring model scoring, all methods use the same judge, prompt, and decoding configuration. The rubric rewards during training are provided by a Qwen3-32B grader. The online rewards of SetGRPO come from the first grader scoring; the second grader scoring is computed only for pre-frozen response groups, used for calibration and excluded from the training coefficients of online groups. 
% The optimizer, batch size, learning rate, calibration set size, candidate thresholds, and hardware configuration are given in Appendix~\ref{app:implementation_details}.
% 每个问题生成 $K=8$ 个rollout回答，最大 response length 为 4096 tokens。各方法使用相同的训练数据、训练步数和生成预算，并在训练结束后使用相同的评测协议。需要模型评分的 benchmark 对所有方法使用相同的 judge、prompt 和 decoding configuration。训练阶段的 rubric reward 由 Qwen3-32B grader 提供。SetGRPO 的在线 reward 由第一次 grader 评分 提供；第二次 grader 评分只对预先冻结的回答组评分，用于校准，不参与在线回答组的训练系数计算。优化器、batch size、学习率、校准集规模、候选阈值和硬件配置见 Appendix~\ref{app:implementation_details}。

\subsection{Main Results}
\label{subsec:main_results}

\begin{figure*}[t]
    \centering
    \vspace{-10pt}
    \includegraphics[width=\textwidth]{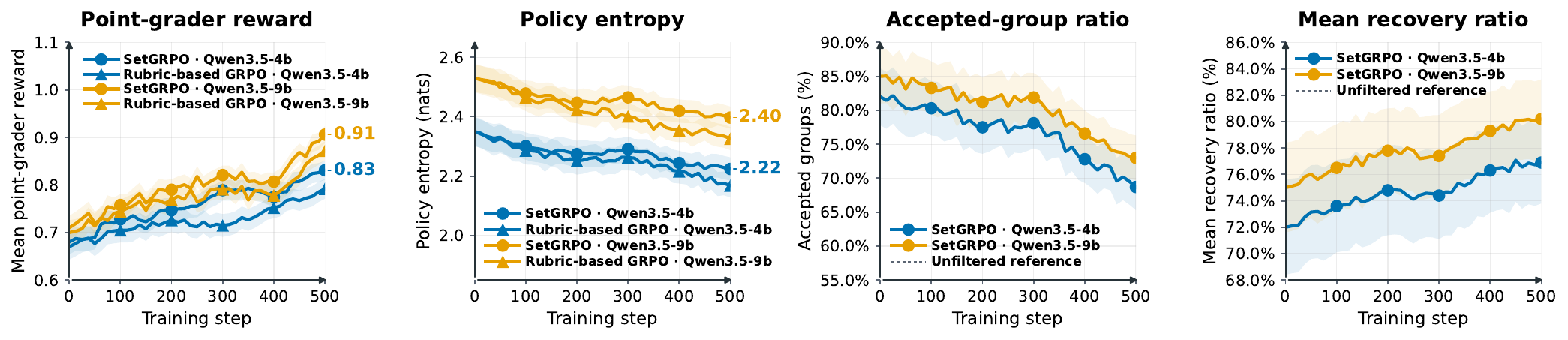}
    \vspace{-10pt}
    \caption{Training dynamics of SetGRPO, including the reward obtained from the first score, policy entropy, accepted-group ratio, and mean recovery ratio over the first 500 training steps.}
    \vspace{-10pt}
    \label{fig:training_curves}
\end{figure*}

\arrayrulecolor{black}

\newcommand{\score}[2]{%
    \ensuremath{#1^{\scriptscriptstyle \pm\,#2}}%
}

\definecolor{setgrpo}{rgb}{0.906,0.953,0.984}

\newcommand{\setcell}[1]{%
    \cellcolor{setgrpo}{#1}%
}

\textbf{Overall performance.}
Table~\ref{tab:main_results} reports results on ten benchmarks across three backbones. SetGRPO achieves the highest Macro Avg. on Qwen3.5-4B and Qwen3.5-9B, reaching 60.26 and 63.51, respectively. On Qwen3.5-9B, it improves over RubricHub by 2.82 points on HealthBench and 2.52 on WritingBench, and over RuscaRL by 4.86 on HealthBench and 4.65 on LLMEval-Med. Relative to RubricHub, its average gains across healthcare, writing, and instruction-following tasks are 1.22, 2.61, and 0.96 points, respectively, with consistent improvements across backbones. These results indicate that SetGRPO improves performance across different backbones and task domains by jointly modeling the training-coefficient shifts caused by score discrepancies and screening for actionable updates in combination with the update geometry of the current policy.
% Table~\ref{tab:main_results} reports results on ten benchmarks for three backbones, with Macro Avg. summarizing the overall performance. Among the four training methods, SetGRPO attains the highest macro average on both Qwen3.5-4B and Qwen3.5-9B, reaching 60.26 and 63.51, respectively. Taking Qwen3.5-9B as an example, compared with RubricHub, SetGRPO improves by 2.82 and 2.52 points on HealthBench and WritingBench, respectively; compared with RuscaRL, the improvement is 4.86 and 4.65 points on HealthBench and LLMEval-Med. Relative to RubricHub, SetGRPO yields an average improvement of 1.22, 2.61, and 0.96 points on the healthcare, writing, and instruction following tasks, respectively, and the overall trend of improvement is consistent across the two backbones. These results indicate that SetGRPO improves performance across different backbones and task domains by jointly modeling the training-coefficient shifts caused by score discrepancies and screening for actionable updates in combination with the update geometry of the current policy.
% Table~\ref{tab:main_results} 报告了三个 backbone 在十个 benchmark 上的结果，Macro Avg. 用于概括整体表现。在四种训练方法中，SetGRPO 在 Qwen3.5-4B 和 Qwen3.5-9B 上均取得最高的 Macro Avg.，分别为 60.26 和 63.51。以 Qwen3.5-9B 为例，相较于 RubricHub，SetGRPO 在 HealthBench 和 WritingBench 上分别提高 2.82 和 2.52 分；相较于 RuscaRL，则在 HealthBench 和 LLMEval-Med 上分别提高 4.86 和 4.65 分。相比于 RubricHub，SetGRPO 在医疗、写作和指令遵循任务上的平均提升分别为 1.22、2.61 和 0.96 分，两个 backbone 上的整体提升趋势一致。这些结果表明，SetGRPO 对评分差异引起的训练系数偏移进行联合建模，并结合当前策略的更新几何筛选可执行更新，提升了不同 backbone 和任务领域上的性能。

\begin{table*}[hb]
    \centering
    \caption{Same-backbone comparison between RubricHub and SetGRPO on Qwen3-4B and Qwen3-14B. RubricHub results were reproduced by our implementation using its publicly available training procedure. HB: HealthBench; LEM: LLMEval-Med; MQA: MedQA; MCQA: MedMCQA; WB: WritingBench; CW: Creative Writing v3; IFE: IFEval; IFB: IFBench; IB: InfoBench; and MIF: MulDimIF.}
    \label{tab:same_backbone_ruscar}
    \renewcommand{\arraystretch}{1.08}
    \setlength{\tabcolsep}{2.5pt}
    \scriptsize
    \resizebox{0.95\textwidth}{!}{%
    \begin{tabular}{@{}llccccccccccc@{}}
    \toprule
    \multirow{2}{*}{\textbf{Backbone}}
    & \multicolumn{1}{c}{\multirow{2}{*}{\textbf{Method}}}
    & \multicolumn{4}{c}{\textbf{Medical}}
    & \multicolumn{2}{c}{\textbf{Writing}}
    & \multicolumn{4}{c}{\textbf{Instruction Following}}
    & \multirow{2}{*}{\textbf{Macro Avg.}} \\
    \cmidrule(lr){3-6}\cmidrule(lr){7-8}\cmidrule(lr){9-12}
    & & \textbf{HB} & \textbf{LEM} & \textbf{MQA} & \textbf{MCQA} & \textbf{WB} & \textbf{CW} & \textbf{IFE} & \textbf{IFB} & \textbf{IB} & \textbf{MIF} & \\
    \midrule
    \multirow{2}{*}{Qwen3-4B} & RubricHub & 64.34 & 45.12 & \textbf{78.38} & \textbf{61.88} & 66.51 & 28.47 & 64.21 & 24.08 & \textbf{80.21} & 47.35 & 56.06  \\
    & \cellcolor{setgrpoBlue}\textbf{SetGRPO (Ours)}
    & \cellcolor{setgrpoBlue}\textbf{66.82}
    & \cellcolor{setgrpoBlue}\textbf{47.57}
    & \cellcolor{setgrpoBlue}77.91
    & \cellcolor{setgrpoBlue}61.42
    & \cellcolor{setgrpoBlue}\textbf{68.97}
    & \cellcolor{setgrpoBlue}\textbf{30.18}
    & \cellcolor{setgrpoBlue}\textbf{66.83}
    & \cellcolor{setgrpoBlue}\textbf{26.49}
    & \cellcolor{setgrpoBlue}79.66
    & \cellcolor{setgrpoBlue}\textbf{49.72}
    & \cellcolor{setgrpoBlue}\textbf{57.56} \\
    \midrule
    \multirow{2}{*}{Qwen3-14B} & Rubrichub & 68.44 & 53.36 & \textbf{77.53} & \textbf{64.76} & 70.42 & 32.71 & 69.74 & 25.61 & \textbf{82.88} & 46.79 & 59.12 \\
    & \cellcolor{setgrpoBlue}\textbf{SetGRPO (Ours)}
    & \cellcolor{setgrpoBlue}\textbf{70.91}
    & \cellcolor{setgrpoBlue}\textbf{54.88}
    & \cellcolor{setgrpoBlue}76.94
    & \cellcolor{setgrpoBlue}64.18
    & \cellcolor{setgrpoBlue}\textbf{72.83}
    & \cellcolor{setgrpoBlue}\textbf{34.62}
    & \cellcolor{setgrpoBlue}\textbf{72.21}
    & \cellcolor{setgrpoBlue}\textbf{27.94}
    & \cellcolor{setgrpoBlue}82.17
    & \cellcolor{setgrpoBlue}\textbf{49.18}
    & \cellcolor{setgrpoBlue}\textbf{60.59} \\
    \bottomrule
    \end{tabular}%
    }
\end{table*}

\textbf{Training dynamics.}
Figure~\ref{fig:training_curves} shows how the mean reward from the first scoring, policy entropy, proportion of accepted groups, and mean recovery ratio over accepted groups evolve during training. The mean reward from the first scoring rises steadily in the early phase and saturates later; policy entropy decreases gradually without an abrupt collapse. Compared with unfiltered training, the proportion of accepted groups is lower but remains nonzero in the later phase, and the mean recovery ratio over accepted groups stays within $76\%$--$80\%$. This indicates that screening excludes response groups that fail the projection condition while nonzero training coefficients still enter the optimization. The steady rise of reward indicates that the retained updates bring training progress, and the stable decline of entropy indicates that the policy does not exhibit marked additional shrinkage.
% Figure~\ref{fig:training_curves} 展示训练过程中平均第一次评分得到的 reward、policy entropy、接受组比例和接受组平均恢复比例的变化。平均第一次评分得到的 reward 在训练前期持续上升，后期趋于稳定；policy entropy 缓慢下降，没有出现骤然收缩。相较于未筛选训练，接受组比例有所降低，但在训练后期仍保持非零，接受组平均恢复比例也维持在 $76\%$--$80\%$。这表明筛选排除了未满足投影条件的回答组，同时仍有非零训练系数进入优化。reward 的持续上升表明保留的更新带来了训练进展；entropy 的平稳下降则表明策略没有出现明显的额外收缩。

% 调节一下字体大小和粗度

\subsection{Same-Backbone Comparison}
\label{app:same_backbone_results}

% 本节在 Qwen3-4B 和 Qwen3-14B 上比较 SetGRPO 与 RubricHub，以保证两种方法使用相同 backbone。与正文一致(Detailed in Section~\ref{subsec:main_results})，我们报告十个 benchmark 的结果，并按 Medical、Writing 和 Instruction Following 三个领域进行分析。

% 在 Qwen3-4B 上，SetGRPO 的 Macro Avg. 为 57.56，较 RubricHub 提高 1.50 分。Medical 领域中，SetGRPO 在 HealthBench 和 LLMEval-Med 上分别提高 2.48 和 2.45 分，在 MedQA 和 MedMCQA 上分别相差 0.47 和 0.46 分。也就是说，SetGRPO 的医疗领域增益主要来自开放式医学评测，同时在选择题评测上保持了接近 RubricHub 的表现。其原因在于，SetGRPO 将组内评分差异经过共享归一化后产生的相关训练系数偏移表示为联合系数集合，再结合当前 response-level gradients 判断候选系数是否能够产生与 reference update 一致的策略更新，从而减少评分差异导致的错误或反向更新。

% Writing 领域中，SetGRPO 在 WritingBench 和 Creative Writing v3 上较 RubricHub 分别提高 2.46 和 1.71 分。对于同时包含内容、文体、格式和长度要求的写作任务，以及强调创意质量的开放式写作任务，SetGRPO 通过恢复满足投影条件的可执行训练进度，保留了与写作质量相关的有效评分信号，因此在两项评测上均取得提升。

% Instruction Following 领域中，SetGRPO 在 IFEval、IFBench 和 MulDimIF 上较 RubricHub 分别提高 2.62、2.41 和 2.37 分，在 InfoBench 上相差 0.55 分。前三项评测的提升表明，SetGRPO 对显式约束、未见约束和多维约束的遵循能力均有改善。对于需要同时满足多个要求的指令，方法先依据联合系数集合和策略更新空间中的投影条件筛选可执行更新，再通过独立风险校准选择统一阈值，减少方向反转更新对多约束行为的破坏。

% 在 Qwen3-14B 上，SetGRPO 的 Macro Avg. 较 RubricHub 提高 1.47 分。两个 backbone 分别完成联合系数集合和执行阈值的校准后，SetGRPO 仍取得一致的整体提升，说明性能增益来自方法对评分差异、策略更新和风险筛选的处理，而不依赖某一个特定 backbone。总体而言，SetGRPO 通过联合表示组内评分差异造成的训练系数偏移，结合当前策略的 response-level gradients 筛选并恢复可执行训练进度，再利用风险校准确定统一执行阈值，从而在不同 backbone 和任务领域上提升最终 benchmark 性能。

In this section, we compare SetGRPO with RubricHub on Qwen3-4B and Qwen3-14B using the same backbone for each comparison. All other experimental settings follow those used for the Qwen3.5 and Llama3.2 experiments, including the data splits, rollout group size, decoding configuration, training steps, generation budget, grader configuration, and evaluation protocol. The backbone-specific implementation details are summarized separately in Tables~\ref{tab:qwen3_training_config} and~\ref{tab:qwen3_calibration_records}. Consistent with the main text (detailed in Section~\ref{subsec:main_results}), we report results on ten benchmarks and analyze them across the Medical, Writing, and Instruction Following domains.

\begin{wraptable}{r}{0.48\textwidth}
\centering
\caption{Training and rollout configuration for Qwen3-4B and Qwen3-14B.}
\label{tab:qwen3_training_config}
\renewcommand{\arraystretch}{1.05}
\setlength{\tabcolsep}{2.5pt}
\scriptsize
\begin{tabular}{@{}lcc@{}}
\toprule
\textbf{Configuration}
& \textbf{Qwen3-4B}
& \textbf{Qwen3-14B} \\
\midrule
Rollout group size $K$ & 8 & 8 \\
Max prompt length & 2048 & 2048 \\
Max response length & 4096 & 4096 \\
Batch size & 64 & 64 \\
Mini-batch size & 32 & 32 \\
Optimizer & AdamW & AdamW \\
Learning rate & $10^{-6}$ & $10^{-6}$ \\
Weight decay & 0.01 & 0.01 \\
KL coefficient & 0.001 & 0.001 \\
Rollout decoding ($T$ / top-$p$ / top-$k$)
& 1.0 / 1.0 / 20
& 1.0 / 1.0 / 20 \\
Training steps & 500 & 500 \\
Checkpoint frequency & 10 & 10 \\
\bottomrule
\end{tabular}
\end{wraptable}

On Qwen3-4B, SetGRPO achieves a Macro Avg. of 57.56, improving over RubricHub by 1.50 points. In the Medical domain, SetGRPO improves over RubricHub by 2.48 and 2.45 points on HealthBench and LLMEval-Med, respectively, while the differences on MedQA and MedMCQA are 0.47 and 0.46 points. Thus, the gain in the medical domain mainly comes from open-ended medical evaluation, while performance on multiple-choice evaluation remains close to that of RubricHub. SetGRPO represents the correlated training-coefficient shifts caused by shared normalization of within-group score discrepancies as a joint coefficient set. It then combines this set with the response-level gradients of the current policy to determine whether candidate coefficients induce policy updates consistent with the reference update, thereby reducing erroneous or reversed updates caused by score discrepancies.

\begin{wraptable}{r}{0.52\textwidth}
\vspace{-10pt}
\centering
\caption{Frozen coverage and risk calibration records for Qwen3-4B and Qwen3-14B. ECR denotes empirical conditional risk.}
\label{tab:qwen3_calibration_records}
\vspace{-5pt}
\renewcommand{\arraystretch}{1.05}
\setlength{\tabcolsep}{2pt}
\scriptsize
\begin{tabular}{@{}lrrrrrrrr@{}}
\toprule
Backbone
& Cov.
& Risk
& $Q_{\alpha}$
& $\tau_{\mathrm{sel}}$
& $N$
& $F$
& ECR
& $U$ \\
\midrule
Qwen3-4B
& 27,229 & 27,229 & 1.32 & 0.25
& 19113 & 1047 & 0.0548 & 0.575 \\
Qwen3-14B
& 27,229 & 27,229 & 1.45 & 0.25
& 19137 & 1058 & 0.0554 & 0.580 \\
\bottomrule
\end{tabular}
\end{wraptable}

In the Writing domain, SetGRPO improves over RubricHub by 2.46 and 1.71 points on WritingBench and Creative Writing v3, respectively. For writing tasks that jointly impose requirements on content, style, format, and length, as well as open-ended tasks emphasizing creative quality, SetGRPO retains effective score signals related to writing quality by recovering executable training progress that satisfies the projection conditions. It therefore improves performance on both evaluations.
In the Instruction Following domain, SetGRPO improves over RubricHub by 2.62, 2.41, and 2.37 points on IFEval, IFBench, and MulDimIF, respectively, while the difference on InfoBench is 0.55 points. The gains on the first three evaluations indicate improved adherence to explicit, unseen, and multidimensional constraints. For instructions requiring multiple requirements to be satisfied simultaneously, the method first screens executable updates using the joint coefficient set and the projection conditions in the policy-update space, and then uses independent risk calibration to select a unified threshold. This reduces the damage that direction-reversed updates can cause to multi-constraint behavior.

On Qwen3-14B, SetGRPO improves the Macro Avg. over RubricHub by 1.47 points. After the joint coefficient set and execution threshold are calibrated separately for the two backbones, SetGRPO achieves consistent overall gains. This indicates that the improvement comes from its treatment of score discrepancies, policy updates, and risk screening rather than dependence on a particular backbone. Overall, SetGRPO jointly represents the training-coefficient shifts caused by within-group score discrepancies, uses the response-level gradients of the current policy to screen and recover executable training progress, and applies risk calibration to determine a unified execution threshold, thereby improving benchmark performance across backbones and task domains.

\section{Analysis and Discussion}
\label{sec:analysis_discussion}

% 本节报告接受更新的条件方向反转风险，并分析不同设计变体、组大小、投影维数和额外计算开销对最终结果的影响。
% 本节报告接受更新的条件方向反转风险，并分析不同设计变体和额外计算开销对最终结果的影响。
This section reports the conditional direction-reversal risk of accepted updates and analyzes the effect of different design variants and the additional computational overhead on the final results.

\subsection{Risk Calibration and Retained Updates}
\label{subsec:risk_recovery}

% Embedded figure: keep the risk-calibration plot alongside the discussion.

% \begin{wrapfigure}{r}{0.48\textwidth}
%     \vspace{-10pt}
%     \centering
%     \includegraphics[
%         width=\linewidth
%     ]{figures/fig_02_main_risk_recovery.pdf}
%     \vspace{-15pt}
%     \caption{Risk calibration and retained updates at the selected threshold
%     $\tau_{\mathrm{sel}}$.}
%     \label{fig:risk_recovery}
%     % \vspace{10pt}
% \end{wrapfigure}

% Figure~\ref{fig:risk_recovery} shows the direction reversal rate, the one-sided upper confidence bound, and the cumulative recovery across candidate thresholds. As the screening condition tightens, both the direction reversal rate and its one-sided upper confidence bound decrease overall, but the number of accepted groups and the cumulative recovery also decrease. In the formal training phase, after selecting $\tau_{\mathrm{sel}}=0.25$, Qwen3.5-4B and Qwen3.5-9B accept 19,115/27,229 and 19,142/27,229 response groups, respectively, corresponding to acceptance rates of 70.2\% and 70.3\%. Their empirical direction-reversal risks are 5.49\% and 5.54\%, with one-sided Clopper--Pearson upper confidence bounds of 5.77\% and 5.82\%, both satisfying the target risk $\beta=0.08$. The nonzero update rates remain substantial during formal training. Thus, the selected threshold satisfies the prescribed conditional-risk constraint while retaining a substantial number of nonzero training updates.

\begin{wrapfigure}{r}{0.56\textwidth}
    \vspace{-20pt}
    \centering
    \includegraphics[width=\linewidth]{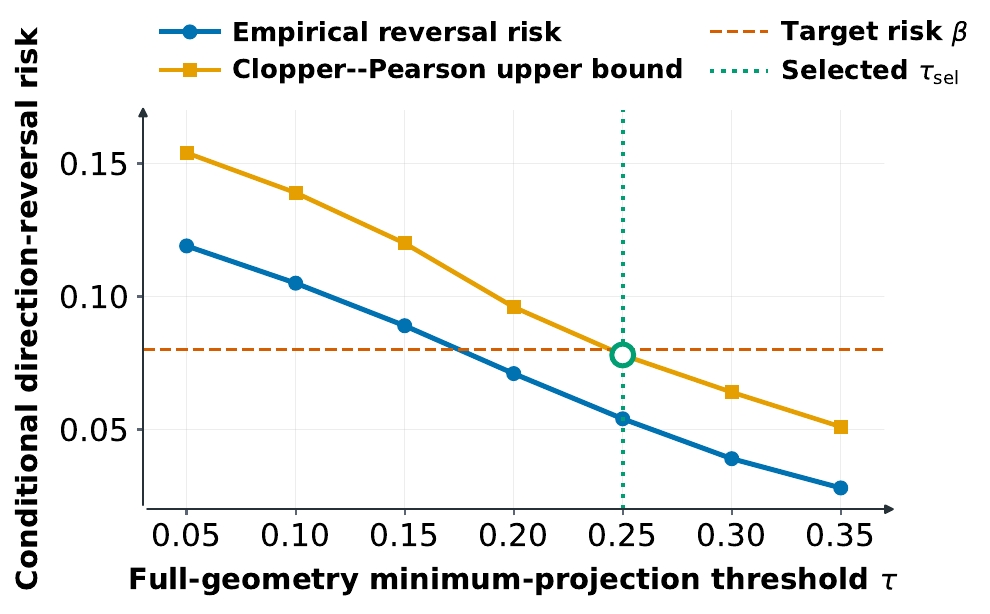}
    \vspace{-18pt}
    \caption{Risk calibration and retained updates at the selected threshold
    $\tau_{\mathrm{sel}}$.}
    \label{fig:risk_recovery}
    \vspace{-15pt}
\end{wrapfigure}

Figure~\ref{fig:risk_recovery} shows the empirical direction-reversal risk and its one-sided Clopper--Pearson upper confidence bound across candidate thresholds, together with the target risk $\beta$ and the selected threshold $\tau_{\mathrm{sel}}=0.25$. As the screening condition tightens, both the empirical risk and its upper confidence bound decrease overall. At the selected threshold, the upper confidence bound remains below $\beta=0.08$. 
% The complete acceptance statistics, cumulative recovery, and formal-training results for each backbone are reported in Appendix~\ref{app:calibration_config}.
% Figure~\ref{fig:risk_recovery} 展示不同候选阈值下的方向反转率、单侧上置信界和累计恢复量。随着筛选条件收紧，方向反转率及其单侧上置信界总体下降，但接受组数量和累计恢复量也随之减少。在正式训练阶段，选定阈值 $\tau_{\mathrm{sel}}=0.25$ 后，Qwen3.5-4B 和 Qwen3.5-9B 分别接受了 19,115/27,229 和 19,142/27,229 个回答组，对应接受率为 70.2\% 和 70.3\%。两者的经验方向反转风险分别为 5.49\% 和 5.54\%，单侧 Clopper--Pearson 上置信界分别为 5.77\% 和 5.82\%，均满足目标风险 $\beta=0.08$。正式训练中的非零更新率不为 0，表明筛选后仍有相当比例的回答组产生更新。因此，所选阈值在控制接受更新的条件方向反转风险的同时，保留了训练更新。

\subsection{Ablation Studies}
\label{subsec:ablation}

To isolate the effects of joint coefficient modeling, policy-update geometry, coefficient recovery, and threshold selection, we evaluate three variant groups: (a) an isotropic coefficient set instead of the joint set; (b) variants removing the policy-space projection condition, coefficient recovery, or reward-space projection condition; and (c) a fixed threshold instead of risk-calibrated threshold selection. We also include a matched random-discard baseline that retains the same number of groups as SetGRPO to distinguish geometry-based screening from merely reducing updates.

\begin{wraptable}{r}{0.65\textwidth}
    \centering
    % \vspace{-15pt}
    \caption{Ablation study of SetGRPO on Qwen3.5-4B. $\Delta$ denotes the change from Full SetGRPO in each metric's native units; upward arrows indicate the preferred direction. CR is normalized cumulative recovery and NUR is the nonzero update rate. Reversal risk, NUR, and CR lie in $[0,1]$.}
    \vspace{-10pt}
    \label{tab:core_ablations}
    \renewcommand{\arraystretch}{1.08}
    \setlength{\tabcolsep}{2.2pt}
    \scriptsize
    \resizebox{\linewidth}{!}{%
    \begin{tabular}{@{}clcccccccc@{}}
    \toprule
    \multirow{2}{*}{\textbf{D\#}}
    & \multirow{2}{*}{\textbf{Derivative}}
    & \multicolumn{2}{c}{\textbf{Macro Avg.}}
    & \multicolumn{2}{c}{\textbf{Reversal risk}}
    & \multicolumn{2}{c}{\textbf{CR}}
    & \multicolumn{2}{c}{\textbf{NUR}} \\
    \cmidrule(lr){3-4}\cmidrule(lr){5-6}\cmidrule(lr){7-8}\cmidrule(lr){9-10}
    & & $\uparrow$ & $\Delta$ & $\downarrow$ & $\Delta$ & $\uparrow$ & $\Delta$ & $\uparrow$ & $\Delta$ \\
    \midrule
        \multicolumn{10}{@{}l}{\cellcolor{gray!8}\textit{(a) Coefficient-set construction}} \\
1 & Spherical set
& 56.62 & \cellcolor{orange!55}{-3.64}
& 0.096 & \cellcolor{orange!45}{+0.049}
& 0.61 & \cellcolor{orange!35}{-0.19}
& 0.512 & \cellcolor{orange!35}{-0.190} \\

        \multicolumn{10}{@{}l}{\cellcolor{gray!8}\textit{(b) Local geometry and recovery}} \\
        
2 & Identity geometry
& 55.08 & \cellcolor{orange!70}{-5.18}
& 0.118 & \cellcolor{orange!70}{+0.071}
& 0.51 & \cellcolor{orange!55}{-0.29}
& 0.438 & \cellcolor{orange!55}{-0.264} \\

3 & Safe-anchor-only
& 56.94 & \cellcolor{orange!45}{-3.32}
& 0.061 & \cellcolor{orange!20}{+0.014}
& 0.36 & \cellcolor{orange!70}{-0.44}
& 0.318 & \cellcolor{orange!70}{-0.384} \\

4 & Reward-only recovery
& 57.78 & \cellcolor{orange!35}{-2.48}
& 0.103 & \cellcolor{orange!55}{+0.056}
& 0.57 & \cellcolor{orange!45}{-0.23}
& 0.566 & \cellcolor{orange!25}{-0.136} \\

        \multicolumn{10}{@{}l}{\cellcolor{gray!8}\textit{(c) Threshold selection and update control}} \\

5 & Fixed-threshold
& 58.11 & \cellcolor{orange!30}{-2.15}
& 0.089 & \cellcolor{orange!40}{+0.042}
& 0.60 & \cellcolor{orange!35}{-0.20}
& 0.548 & \cellcolor{orange!25}{-0.154} \\

6 & Matched random discard
& 54.83 & \cellcolor{orange!70}{-5.43}
& 0.154 & \cellcolor{orange!70}{+0.107}
& 0.49 & \cellcolor{orange!55}{-0.31}
& 0.702 & 0.000 \\

\midrule
\rowcolor{setgrpoBlue}
\multicolumn{2}{@{}l}{\textbf{SetGRPO (Ours)}}
& \textbf{60.26} & \textbf{0.00}
& \textbf{0.047} & \textbf{0.000}
& \textbf{0.80} & \textbf{0.000}
& \textbf{0.702} & \textbf{0.000} \\
    \bottomrule
    \end{tabular}%
    }
    \vspace{-10pt}
\end{wraptable}

Table~\ref{tab:core_ablations} shows that all variants underperform the full setting (Macro Avg.\ 60.26). Replacing the joint coefficient set with a spherical set reduces the Macro Avg.\ to 56.72 and raises reversal risk to 0.096, indicating that this variant fails to preserve the within-group correlated coefficient shifts. Removing the projection condition of the policy-update space reduces the macro average to 55.18 and raises the risk to 0.118, indicating that screening in coefficient space alone is insufficient to determine the effect of coefficient changes on the current policy update. Keeping only the minimum-norm starting point gives a macro average of 57.04 with the recovery amount dropping to 0.36, indicating that not recovering the training coefficients costs training progress. Removing the minimum projection condition in the reward space gives a macro average of 57.88 with the risk rising to 0.103, indicating that the direction screening in the policy-update space cannot substitute for the positive projection condition in the reward space. The fixed threshold variant reaches 58.21 with a risk 0.042 higher than the full setting, indicating that risk calibration contributes to the selection of the execution threshold. Finally, matched random discard reaches only 54.93 despite a comparable acceptance rate, showing that the gains come from targeted screening rather than simply reducing updates.
% Table~\ref{tab:core_ablations} 显示，所有消融变体的 Macro Avg. 均低于完整设置的 60.26。将联合系数集合替换为统一尺度集合后，Macro Avg. 降至 56.72，方向反转风险为 0.096，说明该变体未能保留组内相关的系数偏移。去除策略更新空间的投影条件后，Macro Avg. 降至 55.18，风险升至 0.118，表明系数空间筛选不足以判断系数变化对当前策略更新的影响。仅保留最小范数起点时，Macro Avg. 为 57.04，恢复量降至 0.36，说明不恢复训练系数会导致损失训练进度。去除 reward-space 最小投影条件后，Macro Avg. 为 57.88，风险升至 0.103，说明策略更新空间中的方向筛选不能替代 reward space 中的正向投影条件。固定阈值变体的 Macro Avg. 为 58.21，风险比完整设置高 0.042，说明风险校准有助于选择执行阈值。Matched random discard 在接受组数量相近时仅取得 54.93，低于完整方法，表明性能提升不只来自减少更新数量。

% \begin{table*}[t]

% \end{table*}

\subsection{Efficiency Comparison}
\label{subsec:eff}

\begin{wraptable}{r}{0.63\textwidth}
    \centering
    \vspace{-10pt}
    \caption{Efficiency comparison between GRPO and SetGRPO.}
        \vspace{-10pt}
    \label{tab:efficiency_audit}
    \renewcommand{\arraystretch}{1.08}
    \setlength{\tabcolsep}{2pt}
    \small
    \resizebox{\linewidth}{!}{%
    \begin{tabular}{@{}lcccccccc@{}}
    \toprule
    \multirow{2}{*}{\textbf{Method}}
    & \multicolumn{5}{c}{\textbf{Training-time breakdown}}
    & \multicolumn{2}{c}{\textbf{Resource usage}}
    & \multirow{2}{*}{\makecell[c]{\textbf{10-Benchmark}\\\textbf{Macro Avg.}}} \\
    \cmidrule(lr){2-6}\cmidrule(lr){7-8}
    & \makecell[c]{Single-step\\time (s)}
    & \makecell[c]{Rollout\\time (s)}
    & \makecell[c]{Grader\\time (s)}
    & \makecell[c]{Jacobian\\time (s)}
    & \makecell[c]{Coefficient\\recovery time (s)}
    & \makecell[c]{Peak VRAM\\(GB)}
    & \makecell[c]{Cumulative\\GPU-hours}
    & \\
    \midrule
    GRPO & 82 & 48 & 18 & 0 & 0 & 61 & 22.8 & 55.41 \\
    \rowcolor{setgrpoBlue}
    \textbf{SetGRPO} & 103 & 48 & 18 & 13 & 7 & 67 & 28.6 & 60.26 \\
    \bottomrule
    \end{tabular}%
    }
    \vspace{-5pt}
\end{wraptable}

In Table~\ref{tab:efficiency_audit}, we compare GRPO and SetGRPO in computational cost and benchmark performance. SetGRPO increases per-step training time from 82 to 103 s, peak VRAM from 61 to 67 GB, and cumulative GPU hours from 22.8 to 28.6, corresponding to increases of 25.6\%, 9.8\%, and 25.4\%, while rollout and grader times remain unchanged at 48 s and 18 s. The overhead mainly comes from Jacobian computation for coefficient-set calibration and per-group search for the largest projection-safe recovery ratio. In return, the ten-benchmark Macro Avg.\ increases from 55.41 to 60.26 (+4.75), indicating improved performance at the cost of additional computation.

% Table~\ref{tab:efficiency_audit} compares the original GRPO and SetGRPO in training time, memory footprint, GPU hours, and the macro average over the ten benchmarks. The per-step training time of SetGRPO increases from 82 s to 103 s, the peak VRAM from 61 GB to 67 GB, and the cumulative GPU hours from 22.8 to 28.6, corresponding to increases of 25.6\%, 9.8\%, and 25.4\%. The additional time is spent on the computation of the Jacobian of the training mapping and on the per-group solution of the recovery ratio, while the rollout and grader times remain 48 s and 18 s. The extra computation therefore mainly comes from the Jacobian computation required for calibrating the coefficient set and from determining, within the preset set of candidate ratios, the largest recovery ratio satisfying the projection condition. At the same time, the macro average rises from 55.52 to 60.36, an increase of 4.84 points. Under this experimental setup, SetGRPO attains higher benchmark performance at the cost of additional computational overhead.
% 表~\ref{tab:efficiency_audit} 比较了原始 GRPO 与 SetGRPO 的训练时间、显存占用、GPU-hours 及十个 benchmark 的 Macro Avg.。SetGRPO 的单步训练时间由 82 s 增至 103 s，峰值显存由 61 GB 增至 67 GB，累计 GPU-hours 由 22.8 增至 28.6，分别增加 25.6\%、9.8\% 和 25.4\%。额外时间主要用于训练映射 Jacobian 的计算和逐组恢复比例的求解；rollout 和 grader 时间仍分别为 48 s 和 18 s。因此，额外计算主要来自系数集合校准所需的 Jacobian 计算，以及在预设候选比例集合中确定满足投影条件的最大恢复比例。与此同时，Macro Avg. 从 55.52 提升至 60.36，增加 4.84 分。在本实验设置下，SetGRPO 以增加计算开销为代价，取得了更高的 benchmark 性能。

\begin{figure*}[t]
    \centering
    \begin{subfigure}[t]{0.55\textwidth}
        \centering
        \includegraphics[width=\linewidth]{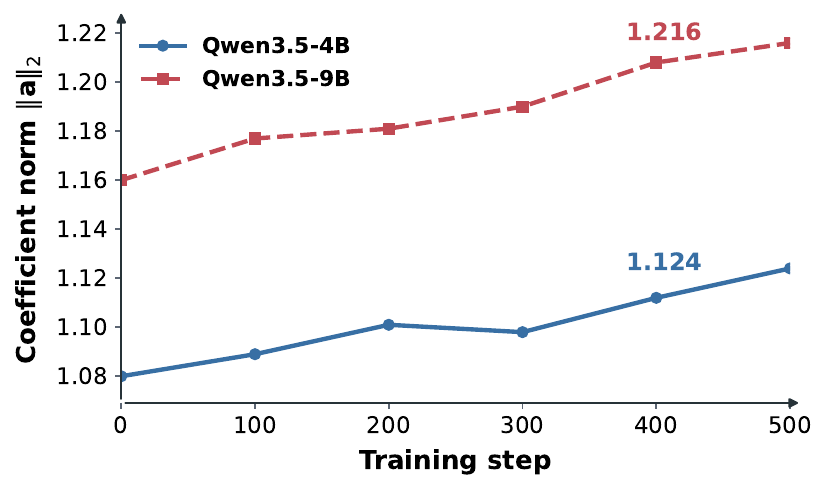}
        \vspace{-20pt}
        \caption{Coefficient-norm trajectory.}
        \label{fig:coefficient_norm_a}
    \end{subfigure}
    \hfill
    \begin{subfigure}[t]{0.40\textwidth}
        \centering
        \includegraphics[width=\linewidth]{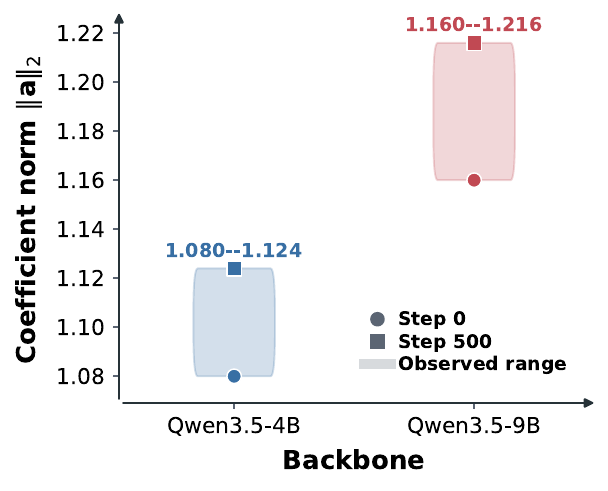}
        \vspace{-20pt}
        \caption{Observed norm range.}
        \label{fig:coefficient_norm_b}
    \end{subfigure}
    \caption{Coefficient-norm dynamics during training. The left panel shows
    the coefficient-norm trajectories of the two backbones, while the right
    panel summarizes the observed norm range between training steps 0 and 500.
    The bounded ranges indicate that coefficient recovery does not lead to
    sustained numerical growth.}
    \label{fig:coefficient_norm}
\end{figure*}

\subsection{Recovery and Selection Diagnostics}
\label{app:training_diagnostics}

% 我们在 Figure~\ref{fig:training_curves} 中报告了主要训练曲线。本节进一步补充 coefficient norm、full-space/reward-space minimum projection 和 cumulative recovery，分别考察恢复系数的规模、联合系数集合上的投影条件以及训练过程中保留的恢复量。

% \textbf{1) Coefficient-Norm Dynamics.}
% Figure~\ref{fig:coefficient_norm} 展示训练过程中恢复系数的范数变化。系数范数整体保持稳定，并随策略状态适度变化，没有出现持续增大的趋势。该结果与方法中由联合系数集合确定恢复范围、并沿最小更新范数锚点到 point-grader 训练系数的恢复路径相一致，说明性能提升来自对可执行恢复更新的筛选与保留，而不是持续放大训练系数。

% \textbf{2) Minimum-Projection Diagnostics.}
% Figure~\ref{fig:projection_diagnostics} 展示 full-space minimum projection 和 reward-space minimum projection。接受更新的 full-space minimum projection 保持在选定阈值 $\tau_{\mathrm{sel}}$ 以上，reward-space minimum projection 保持为正；未产生恢复更新的回答组至少有一项条件未满足。该结果表明，筛选同时检查联合系数集合 $\mathcal C_x$ 下的策略更新投影和 reward-space 投影，使保留的更新与参考训练系数 $\mathbf a_x^\star$ 所诱导的局部更新保持一致，并维持 reward space 中的正投影。

% \textbf{3) Cumulative Recovery.}
% Figure~\ref{fig:cumulative_recovery} 展示累计恢复量随训练步数的变化。累计恢复量持续增加，但增长速度随训练阶段变化，反映了不同阶段可满足恢复条件的回答组数量及其最大恢复比例存在差异。该曲线表明，正式训练过程中持续有回答组满足恢复条件并写回非零恢复系数，因此保留的训练进度并非集中在少数训练步。

We report the main training curves in Figure~\ref{fig:training_curves}. This section further reports the coefficient norm, full-space and reward-space minimum projections, and cumulative recovery to examine the scale of recovered coefficients, the projection conditions over the joint coefficient set, and the recovery retained during training.

\begin{figure*}[t]
    \centering

    \begin{subfigure}[t]{0.48\textwidth}
        \centering
        \includegraphics[
            width=\linewidth
        ]{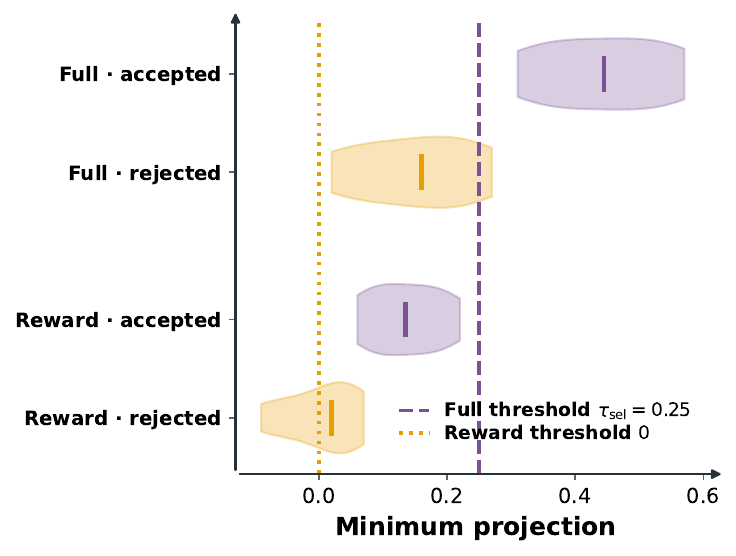}
        \caption{Projection distributions.}
        \label{fig:projection_diagnostics_distributions}
    \end{subfigure}
    \hfill
    \begin{subfigure}[t]{0.48\textwidth}
        \centering
        \includegraphics[
            width=\linewidth
        ]{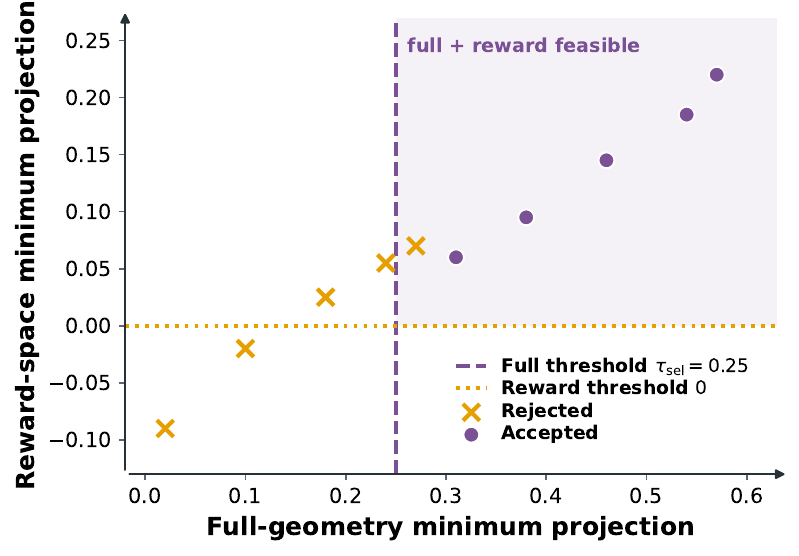}
        \caption{Joint projection constraints.}
        \label{fig:projection_diagnostics_joint}
    \end{subfigure}

    \caption{Projection diagnostics for accepted and rejected updates.
    The left panel shows the distributions of full-geometry and reward-space
    minimum projections. The right panel shows their joint constraints; the
    shaded upper-right region satisfies both the full-geometry threshold
    $\tau_{\mathrm{sel}}$ and the positive reward-space projection condition.}
    \label{fig:projection_diagnostics}
\end{figure*}

\textbf{1) Coefficient-Norm Dynamics.}
Figure~\ref{fig:coefficient_norm} shows the norm of the recovered coefficients throughout training. The coefficient norm remains generally stable and changes moderately with the policy state, without a sustained upward trend. This result is consistent with the recovery range determined by the joint coefficient set and the recovery path from the minimum-update-norm anchor to the point-grader training coefficients. It indicates that the performance gain comes from screening and retaining executable recovery updates rather than continuously enlarging the training coefficients.

\begin{wrapfigure}{r}{0.6\textwidth}
   \centering
   \vspace{-10pt}
   \includegraphics[
       width=\linewidth
   ]{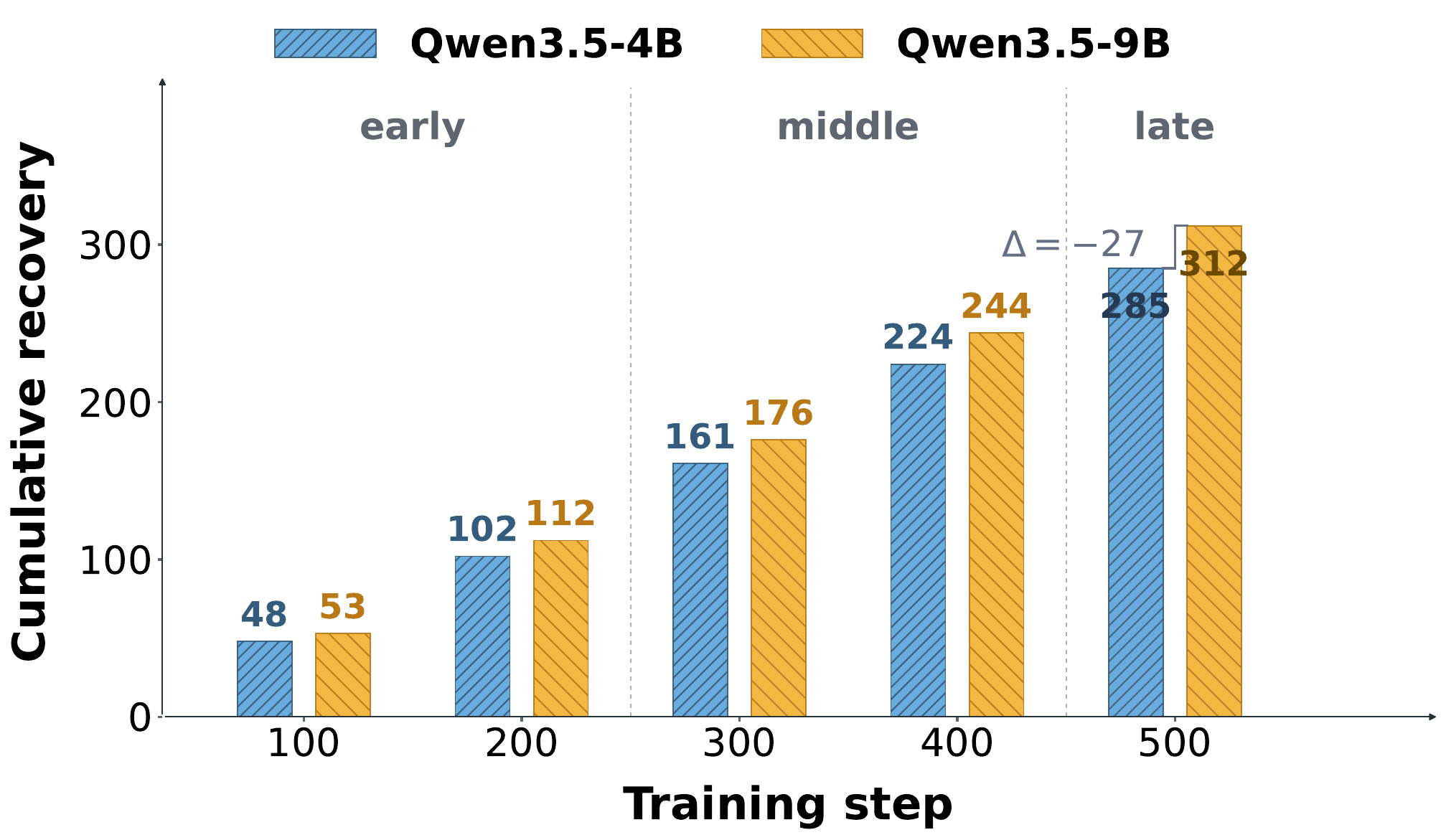}
   \vspace{-20pt}
   \caption{Cumulative recovery over the first 500 training steps for
   Qwen3.5-4B and Qwen3.5-9B. Shaded regions distinguish the early, middle, and late stages; the bracket reports the final cumulative-recovery gap.}
   \label{fig:cumulative_recovery}
   \vspace{-8pt}
\end{wrapfigure}

\textbf{2) Minimum-Projection Diagnostics.}
Figure~\ref{fig:projection_diagnostics} shows the full-space and reward-space minimum projections. For accepted updates, the full-space minimum projection remains above the selected threshold $\tau_{\mathrm{sel}}$, while the reward-space minimum projection remains positive. Response groups that produce no recovery update fail at least one of these conditions. This result shows that screening jointly checks the policy-update projection under the joint coefficient set $\mathcal C_x$ and the reward-space projection, thereby keeping retained updates consistent with the local updates induced by the reference training coefficients $\mathbf a_x^\star$ while maintaining a positive projection in reward space.

\textbf{3) Cumulative Recovery.}
Figure~\ref{fig:cumulative_recovery} shows cumulative recovery as a function of training steps. Cumulative recovery continues to increase, but its growth rate varies across training stages, reflecting differences in the number of response groups satisfying the recovery conditions and in their maximum recovery ratios. The curve shows that response groups continue to satisfy the recovery conditions and write back nonzero recovered coefficients throughout formal training. Thus, the retained training progress is not concentrated in only a few training steps.

\begin{figure*}[tb]
    \centering
    \includegraphics[
        width=0.8\textwidth]{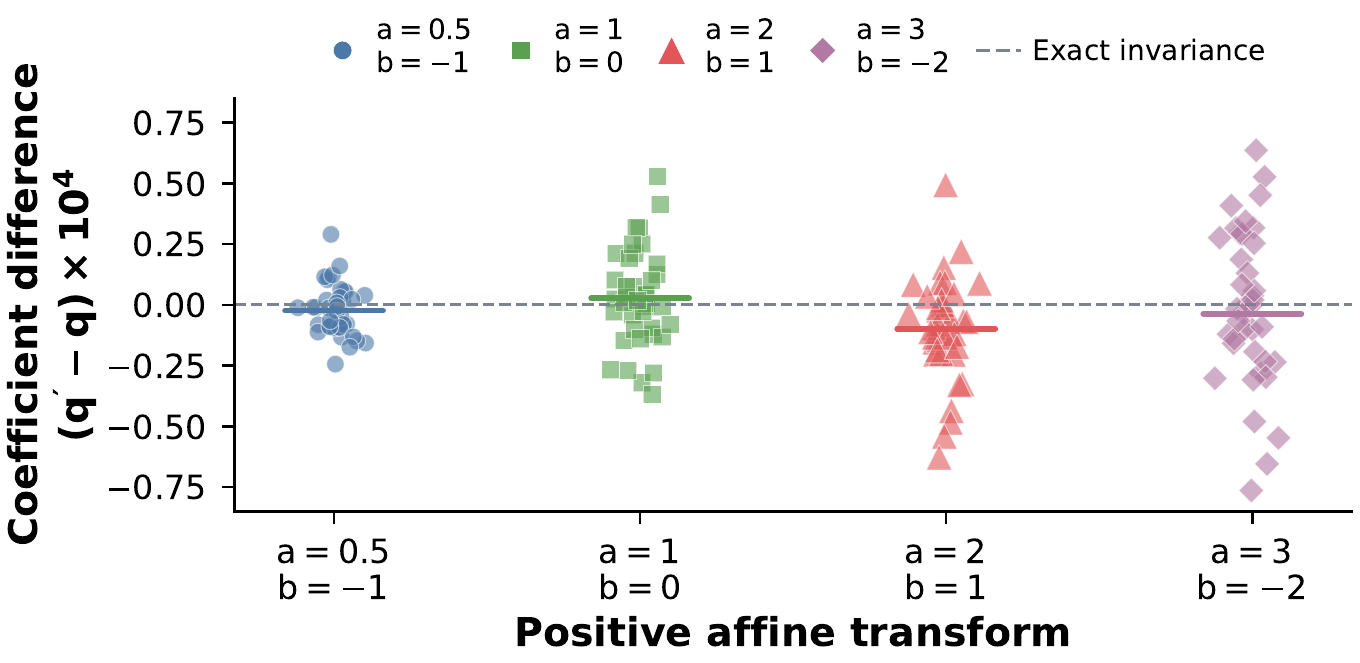}
    \caption{Positive-affine invariance under frozen-policy control. The left
    plot reports coefficient differences between the original and transformed
    standardized coefficients. The right table summarizes coefficient cosine,
    update cosine, decision flips, and the absolute recovery-ratio difference
    for each transform $\mathbf r'_x=a\mathbf r_x+b\mathbf 1_K$ with $a>0$.}
    \label{fig:affine_invariance}
\end{figure*}

\begin{figure*}[t]
    \centering

    \begin{subfigure}[t]{0.48\textwidth}
        \centering
        \includegraphics[
            width=\linewidth
        ]{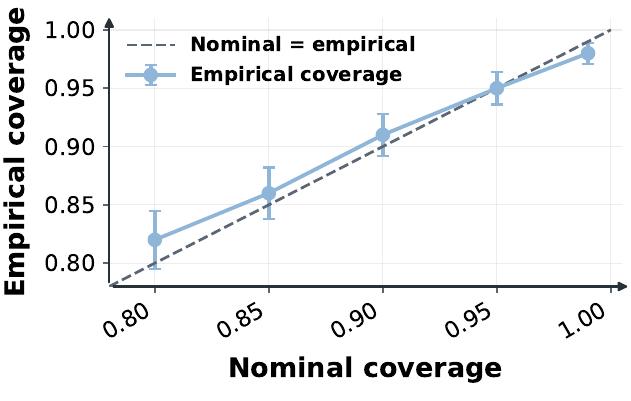}
        \vspace{-20pt}
        \caption{Coverage of the joint coefficient set.}
        \label{fig:coverage_selection_coverage}
    \end{subfigure}
    \hfill
    \begin{subfigure}[t]{0.48\textwidth}
        \centering
        \includegraphics[
            width=\linewidth
        ]{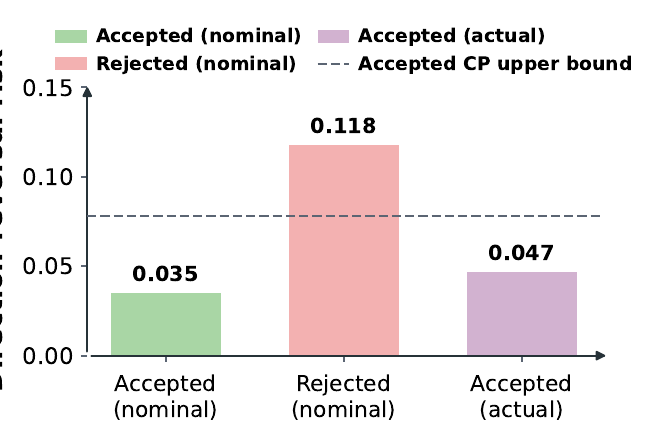}
        \vspace{-20pt}
        \caption{Risk separation at the selected threshold.}
        \label{fig:coverage_selection_risk}
    \end{subfigure}

    \caption{Empirical coverage of the joint coefficient set and
    direction-reversal risk separation at the selected threshold.}
    \label{fig:coverage_selection}
\end{figure*}

\subsection{Affine-Invariance Sanity Check}
\label{app:affine_invariance}

% 训练系数由组内 reward 的中心化与尺度归一化结果决定。为检验这一性质，我们在冻结的回答组和策略快照上构造正仿射变换，
% \begin{equation}
%     \mathbf r'_x=a\mathbf r_x+b\mathbf 1_K,\qquad a>0.
% \end{equation}
% 由于正仿射变换不改变组内标准化 reward，变换前后的训练系数及其联合系数集合应保持一致。我们在相同的 response-level gradients、$\mathbf G_x^q$ 和 $\mathbf u_x^q(\mathbf a)$ 下，固定校准得到的风险规则，比较两种 reward 表示所产生的局部更新、最大恢复比例和进入训练的回答组。

Training coefficients are determined by the centering and scale normalization of rewards within each group. To test this property, we apply a positive-affine transformation to frozen response groups and policy snapshots:
\begin{equation}
   \mathbf r'_x=a\mathbf r_x+b\mathbf 1_K,\qquad a>0.
\end{equation}
Because a positive-affine transformation does not change within-group standardized rewards, the training coefficients and joint coefficient sets before and after transformation should remain unchanged. Under the same response-level gradients, $\mathbf G_x^q$, and $\mathbf u_x^q(\mathbf a)$, with the calibrated risk rule fixed, we compare the local updates, maximum recovery ratios, and response groups entering training under the two reward representations.

% Figure~\ref{fig:affine_invariance} 给出了四种正仿射变换下的系数残差 $(\mathbf a'_x-\mathbf a_x)\times10^4$。各组点云均围绕零线分布，组内均值接近零，说明变换前后的标准化训练系数仅存在数值容差范围内的差异。Table~\ref{tab:affine_invariance} 进一步报告了对应的方向和恢复指标：coefficient cosine 和 update cosine 分别保持在 $0.9997$--$1.0000$ 和 $0.9995$--$1.0000$，四种变换均未产生 decision flip，恢复比例差异最大仅为 $0.8\times10^{-3}$。因此，正仿射变换前后的训练系数、局部更新和恢复比例均未出现系统性变化，说明筛选结果对 reward 的平移和正比例缩放保持稳定。
% 这一结果来自组内标准化对正仿射变换的消除作用。对于 $\mathbf r'_x=a\mathbf r_x+b\mathbf 1_K$ 且 $a>0$，中心化和尺度归一化后的 reward 保持不变。因此，Eq.~\ref{eq:nominal_coefficient} 得到的训练系数保持不变，基于配对系数残差构造的联合系数集合也不会发生实质变化。在当前策略的 response-level gradients、$\mathbf G_x^q$ 和 $\mathbf u_x^q(\mathbf a)$ 保持不变的情况下，策略更新空间和 reward space 中的更新及其 minimum projection 也保持稳定。沿同一恢复路径计算时，$\eta_x^\star$ 仅出现浮点误差量级的变化。

% 该实验表明，SetGRPO 的训练系数、局部更新、minimum projection 和恢复比例由组内 reward 的相对结构决定。对整组 reward 同时进行平移或正比例缩放，不会系统性改变这些量。相反，若变换不再保持组内 reward 的相对顺序，例如使用负比例变换或对不同回答施加不同的非仿射扰动，则组内标准化结果可能发生变化，进而改变训练系数及后续的 minimum projection 和恢复比例。

\begin{wraptable}{r}{0.6\textwidth}
    \centering
    \vspace{-8pt}
    \caption{Positive-affine invariance diagnostics.}
    \label{tab:affine_invariance}
    \setlength{\tabcolsep}{1.5pt}
    \scriptsize
    \resizebox{\linewidth}{!}{%
        \begin{tabular}{@{}lcccc@{}}
            \toprule
            Transform
            & Coeff. cosine
            & Update cosine
            & Flip (\%)
            & $|\Delta\eta^\star|$ ($10^{-3}$) \\
            \midrule
            $a=0.5,\ b=-1$ & 0.9998 & 0.9996 & 0.00 & 0.6 \\
            $a=1,\ b=0$    & 1.0000 & 1.0000 & 0.00 & 0.0 \\
            $a=2,\ b=1$    & 0.9999 & 0.9998 & 0.00 & 0.4 \\
            $a=3,\ b=-2$   & 0.9997 & 0.9995 & 0.00 & 0.8 \\
            \bottomrule
        \end{tabular}%
    }
    \vspace{-8pt}
\end{wraptable}

Figure~\ref{fig:affine_invariance} shows the coefficient residuals $(\mathbf a'_x-\mathbf a_x)\times10^4$ under four positive-affine transformations. The point clouds are centered around zero, and the within-group means are close to zero, indicating that the standardized training coefficients differ only within numerical tolerance. Table~\ref{tab:affine_invariance} reports the corresponding directional and recovery metrics. The coefficient cosine and update cosine range from $0.9997$ to $1.0000$ and from $0.9995$ to $1.0000$, respectively. None of the four transformations produces a decision flip, and the largest difference in recovery ratio is only $0.8\times10^{-3}$. Thus, the training coefficients, local updates, and recovery ratios remain free of systematic changes under positive-affine transformations, indicating that the screening results are stable to reward translation and positive rescaling.

This invariance follows from the elimination of positive-affine transformations by within-group normalization. For $\mathbf r'_x=a\mathbf r_x+b\mathbf 1_K$ with $a>0$, the centered and scale-normalized rewards remain unchanged. Consequently, the training coefficients obtained from Eq.~\ref{eq:nominal_coefficient} remain unchanged, and the joint coefficient set constructed from the paired coefficient residuals does not change materially. With the response-level gradients, $\mathbf G_x^q$, and $\mathbf u_x^q(\mathbf a)$ of the current policy fixed, the updates and their minimum projections in the policy-update space and reward space also remain stable. Along the same recovery path, $\eta_x^\star$ changes only at the level of floating-point error.

This experiment shows that the training coefficients, local updates, minimum projections, and recovery ratios in SetGRPO are determined by the relative structure of within-group rewards. Translating or positively rescaling all rewards in a group does not systematically change these quantities. In contrast, transformations that do not preserve the within-group reward ordering, such as negative rescaling or response-specific non-affine perturbations, may alter the normalized rewards and consequently change the training coefficients, minimum projections, and recovery ratios.

\begin{figure}[t]
    \centering
    \includegraphics[width=\linewidth]{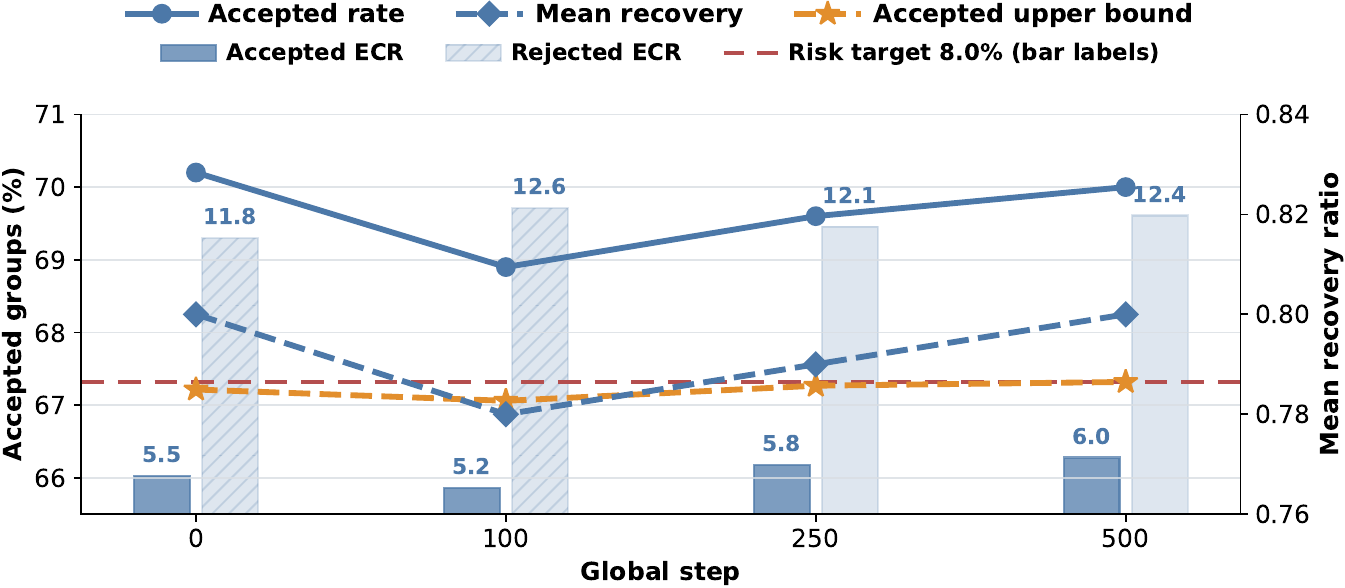}
    \caption{Checkpoint-wise validation of the fixed SRGO rule under policy drift on
    Qwen3.5-4B. Curves show the accepted-group rate, mean recovery ratio, and
    accepted-group upper confidence bound; bars show empirical direction-reversal
    rates for accepted and rejected groups. The red dashed line denotes
    $\beta=0.08$.}
    \label{fig:policy_drift_validation}
\end{figure}

\subsection{Empirical Coverage and Risk Selection}
\label{app:coverage_selection}

Figure~\ref{fig:coverage_selection} reports the coverage performance of the joint coefficient set and the post-screening direction-reversal risk. For each predefined nominal coverage level, we compute $Q_\alpha$ on the coverage calibration groups and compare nominal and empirical coverage on independent risk calibration groups. Empirical coverage varies with the target level and remains close to it overall. Under the final setting, both nominal and empirical coverage are $0.95$, indicating that the joint coefficient set constructed from coefficient residuals standardized by $\mathbf S_x^{\mathrm{coef}}$ covers the reference training coefficients. This result is consistent with Eq.~\ref{eq:jrsc_set}: when the calibration groups and the response groups to be covered satisfy exchangeability, $Q_\alpha$ provides the corresponding finite-sample marginal coverage guarantee.

% 字体调大调粗

% Figure~\ref{fig:coverage_selection}(b) 比较 $\tau_{\mathrm{sel}}$ 下进入训练的回答组与未进入训练的回答组中，两次评分得到的训练系数所诱导的局部更新之间的方向关系。根据 Eq.~\ref{eq:srgo_selection}，方向反转事件 $Z_{x,\tau}$ 及其条件风险 $R_\tau$ 仅对存在非零恢复更新的回答组定义。对于未进入训练的回答组，我们仅报告由两次评分得到的训练系数所诱导的局部更新之间的对照指标，
% \begin{equation}
%     Z_x^{\mathrm{nom}}=
%     \mathbb I\!\left[\left\langle
%     \mathbf u_x^{\mathrm{full}}(\hat{\mathbf a}_x),
%     \mathbf u_x^{\mathrm{full}}(\mathbf a_x^\star)
%     \right\rangle<0\right].
% \end{equation}

% 在图中，进入训练的回答组满足 $\Pr(Z_x^{\mathrm{nom}}=1\mid A_{x,\tau_{\mathrm{sel}}}=1)=0.035$，未进入训练的回答组对应的比例为 $\Pr(Z_x^{\mathrm{nom}}=1\mid A_{x,\tau_{\mathrm{sel}}}=0)=0.118$。对于真正进入训练的回答组，经验方向反转风险为 $0.047$，其单侧 Clopper--Pearson 上置信界为 $U_{\tau_{\mathrm{sel}}}=0.078$，满足 $U_{\tau_{\mathrm{sel}}}\leq\beta$。这表明，筛选后的训练回答组具有更低的方向反转比例。其原因是 Eq.~\ref{eq:ggsp_recovery} 只保留存在非零可行恢复比例、满足 $\Psi_x^{\mathrm{full}}\geq\tau_{\mathrm{sel}}$ 且 $\Psi_x^{\mathrm{reward}}>0$的回答组：前者要求候选更新在联合系数集合下保持足够的策略更新投影，后者要求其在 reward space 中保持正投影。因此，风险校准后的筛选同时改变了进入训练的回答组组成，并将接受更新的条件方向反转风险控制在目标范围内。

Figure~\ref{fig:coverage_selection}(b) compares the directional relationship between local updates induced by the training coefficients from the two scores for response groups entering and not entering training under $\tau_{\mathrm{sel}}$. According to Eq.~\ref{eq:srgo_selection}, the direction-reversal event $Z_{x,\tau}$ and its conditional risk $R_\tau$ are defined only for response groups with nonzero recovery updates. For groups not entering training, we report only the comparison indicator between the local updates induced by the training coefficients from the two scores:
\begin{equation}
   Z_x^{\mathrm{nom}}=
   \mathbb I!\left[
   \left\langle
   \mathbf u_x^{\mathrm{full}}(\hat{\mathbf a}_x),
   \mathbf u_x^{\mathrm{full}}(\mathbf a_x^\star)
   \right\rangle<0
   \right].
\end{equation}

In the figure, response groups entering training satisfy $\Pr(Z_x^{\mathrm{nom}}=1\mid A_{x,\tau_{\mathrm{sel}}}=1)=0.035$, whereas the corresponding proportion is $\Pr(Z_x^{\mathrm{nom}}=1\mid A_{x,\tau_{\mathrm{sel}}}=0)=0.118$ for groups not entering training. For groups that actually enter training, the empirical direction-reversal risk is $0.047$, with a one-sided Clopper--Pearson upper confidence bound of $U_{\tau_{\mathrm{sel}}}=0.078$, satisfying $U_{\tau_{\mathrm{sel}}}\leq\beta$. Thus, the screened training groups exhibit a lower direction-reversal rate. This is because Eq.~\ref{eq:ggsp_recovery} retains only groups with a nonzero feasible recovery ratio that satisfy $\Psi_x^{\mathrm{full}}\geq\tau_{\mathrm{sel}}$ and $\Psi_x^{\mathrm{reward}}>0$. The first condition requires the candidate update to maintain sufficient policy-update projection under the joint coefficient set, while the second requires a positive projection in reward space. Therefore, risk-calibrated screening changes the composition of groups entering training and keeps the acceptance-conditioned direction-reversal risk within the target range.

\subsection{Checkpoint-wise Validation under Policy Drift}
\label{app:policy_drift_validation}

% Figure~\ref{fig:policy_drift_validation} 展示了固定校准结果在训练轨迹上的检查点验证。我们在 calibration 阶段确定 $Q_{\alpha}$ 和 $\tau_{\mathrm{sel}}$，随后在 global step $t\in\{0,100,250,500\}$ 使用同一回答组重新执行 SRGO。对于每个检查点，仅根据当前策略的 response-level gradients 计算投影后的局部更新、$\Psi_x^{\mathrm{full}}$、$\Psi_x^{\mathrm{reward}}$ 及恢复比例；$Q_{\alpha}$ 和 $\tau_{\mathrm{sel}}$ 不随检查点改变。

% 随着训练推进，当前策略的局部更新几何发生变化，但筛选行为没有出现突变。图~\ref{fig:policy_drift_validation}(a) 中，接受组比例从初始检查点到中期略有下降，随后趋于稳定；图~\ref{fig:policy_drift_validation}(b) 中，接受组的平均恢复比例始终为正，说明策略变化并未使可行恢复路径整体消失。与此同时，接受组的经验条件方向反转率保持在较低水平，其单侧上置信界在各检查点均未出现明显抬升；拒绝组的方向反转率则始终更高。该结果表明，策略漂移主要改变了回答组落在联合系数集合中的位置，而没有破坏由 $\Psi_x^{\mathrm{full}}\geq\tau_{\mathrm{sel}}$ 和 $\Psi_x^{\mathrm{reward}}>0$ 定义的筛选关系。

% 该结果与校准阶段确定的联合系数集合和风险选择阈值相一致。正式训练中，$Q_{\alpha}$ 与 $\tau_{\mathrm{sel}}$ 保持固定；各检查点只根据当时的局部梯度重新计算 $\Psi_x^{\mathrm{full}}$、$\Psi_x^{\mathrm{reward}}$ 及恢复比例。因此，检查点间的差异反映的是局部更新几何变化对恢复比例的影响，而不是校准量的重新估计。满足 $\Psi_x^{\mathrm{full}}\geq\tau_{\mathrm{sel}}$ 且 $\Psi_x^{\mathrm{reward}}>0$ 的回答组保留非零恢复系数，其余回答组停止恢复并写回零系数。该实验结果显示，策略漂移会改变回答组的投影值和恢复比例，但接受与拒绝仍由同一组投影条件判定。接受组保持较低的方向反转风险和正的恢复比例，说明该筛选条件在训练轨迹上仍能区分可恢复更新与不满足条件的更新。

Figure~\ref{fig:policy_drift_validation} presents checkpoint validation of the fixed calibration results along the training trajectory. During calibration, we determine $Q_{\alpha}$ and $\tau_{\mathrm{sel}}$, and then rerun SRGO on the same response groups at global steps $t\in\{0,100,250,500\}$. At each checkpoint, the projected local updates, $\Psi_x^{\mathrm{full}}$, $\Psi_x^{\mathrm{reward}}$, and recovery ratios are computed solely from the response-level gradients of the current policy. The values of $Q_{\alpha}$ and $\tau_{\mathrm{sel}}$ remain unchanged across checkpoints.

As training proceeds, the local update geometry of the current policy changes, but the screening behavior does not exhibit abrupt changes. In Figure~\ref{fig:policy_drift_validation}(a), the accepted-group proportion decreases slightly from the initial to the intermediate checkpoint and then stabilizes. In Figure~\ref{fig:policy_drift_validation}(b), the mean recovery ratio of accepted groups remains positive, indicating that policy changes do not eliminate feasible recovery paths. Meanwhile, the empirical conditional direction-reversal rate of accepted groups remains low, and its one-sided upper confidence bound shows no substantial increase at any checkpoint. Rejected groups consistently exhibit higher direction-reversal rates. These results indicate that policy drift mainly changes the positions of response groups relative to the joint coefficient set, without disrupting the screening relation defined by $\Psi_x^{\mathrm{full}}\geq\tau_{\mathrm{sel}}$ and $\Psi_x^{\mathrm{reward}}>0$.

This result is consistent with the joint coefficient set and risk-selection threshold fixed during calibration. During formal training, $Q_{\alpha}$ and $\tau_{\mathrm{sel}}$ remain fixed, while each checkpoint recomputes $\Psi_x^{\mathrm{full}}$, $\Psi_x^{\mathrm{reward}}$, and the recovery ratios using only the local gradients at that checkpoint. Thus, differences across checkpoints reflect the effect of changing local update geometry on recovery ratios rather than re-estimation of the calibration quantities. Response groups satisfying $\Psi_x^{\mathrm{full}}\geq\tau_{\mathrm{sel}}$ and $\Psi_x^{\mathrm{reward}}>0$ retain nonzero recovered coefficients, whereas the remaining groups stop recovery and write back zero coefficients. The results show that policy drift changes projection values and recovery ratios, while acceptance and rejection continue to be determined by the same projection conditions. Accepted groups retain lower direction-reversal risk and positive recovery ratios, indicating that the screening rule continues to distinguish recoverable updates from updates that fail the conditions throughout the training trajectory.

\section{Representative Cases}
\label{app:representative_cases}

% Figures~\ref{fig:representative_cases_1}--\ref{fig:representative_cases_3} 分别展示 accepted、rejected 和 recovery-infeasible 三类回答组，每类包含三个代表性案例。每个案例列出两次评分、对应的训练系数、策略更新空间与 reward space 中的 minimum projection、恢复比例 $\eta_x^\star$ 及最终决策原因。对于 accepted 案例，以 WritingBench \#4 为例，两个 minimum projection 分别为 0.84 和 0.81，满足 $\Psi_x^{\mathrm{full}}\geq\tau_{\mathrm{sel}}$ 与 $\Psi_x^{\mathrm{reward}}>0$，最大可行恢复比例为 $\eta_x^\star=0.82$。rejected 案例均存在非零候选更新，但未满足至少一项投影条件。例如，HealthBench 920ca4e9 的 $M^{\mathrm{full}}=0.29$，满足 full-space 条件，但 $M^{\mathrm{reward}}=-0.06$，未满足 reward-space 正投影条件，因此恢复比例为 $\eta^\star=0.00$。recovery-infeasible 案例则不存在满足 Eq.~\ref{eq:ggsp_recovery} 的非零可行恢复比例。例如，IFEval \#2035 的 $M^{\mathrm{full}}=0.11$ 和 $M^{\mathrm{reward}}=-0.03$ 均未满足相应条件，因此其恢复状态为 infeasible。三类案例的两次评分没有随决策结果呈现单调变化，而 minimum projection 和最大可行恢复比例直接对应 Eq.~\ref{eq:ggsp_recovery} 的筛选条件。

% 不同案例的决策差异来自相同评分差异在不同 response-level gradients 下产生了不同的局部更新。SetGRPO 将评分所得训练系数的联合不确定性表示为 $\mathcal C_x$，再结合当前策略的 response-level gradients 检查候选更新在策略更新空间和 reward space 中的 minimum projection，并确定各回答组可写回的最大恢复比例。因此，满足投影条件的回答组能够保留非零训练更新，无法满足条件的回答组则不会写回更新。

\begin{figure*}[htbp]
    \centering
    \includegraphics[width=\textwidth]{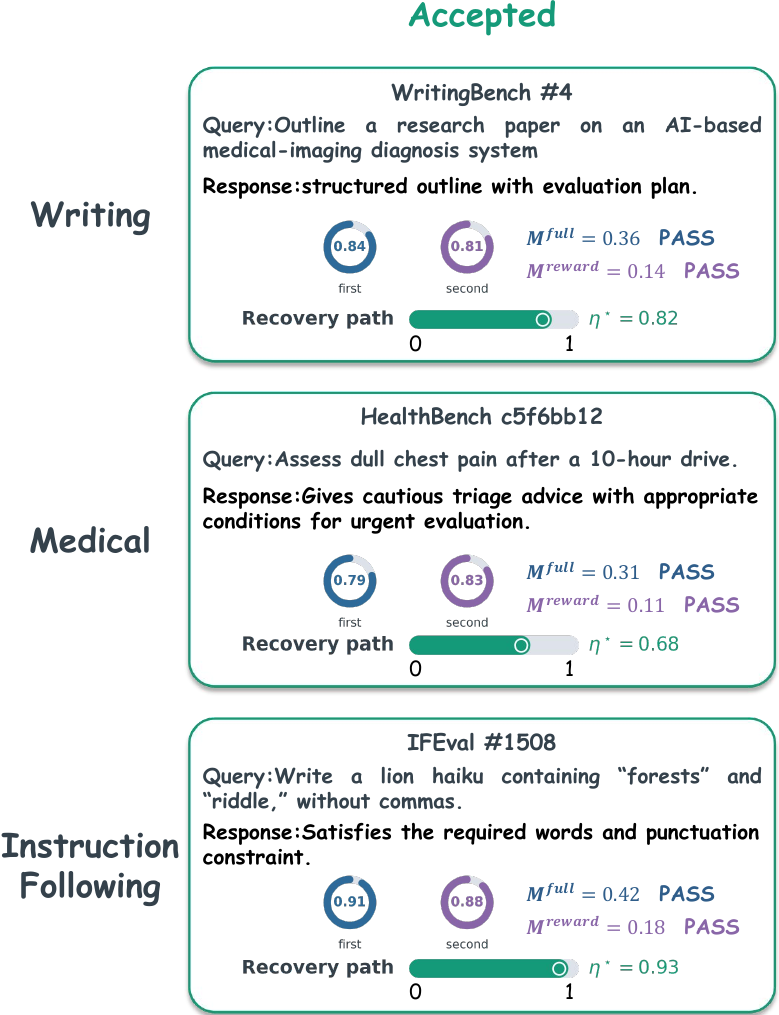}
    \caption{Representative accepted cases.
    Each case shows the anonymized query--response record, point and reference
    scores, full-geometry and reward-space projection checks, the recovery path
    from $\eta=0$ to $\eta=1$, the recovered training coefficient
    $\hat{\mathbf a}_x$, and the final decision.}
    \label{fig:representative_cases_1}
\end{figure*}

\begin{figure*}[htbp]
    \centering
    \includegraphics[width=\textwidth]{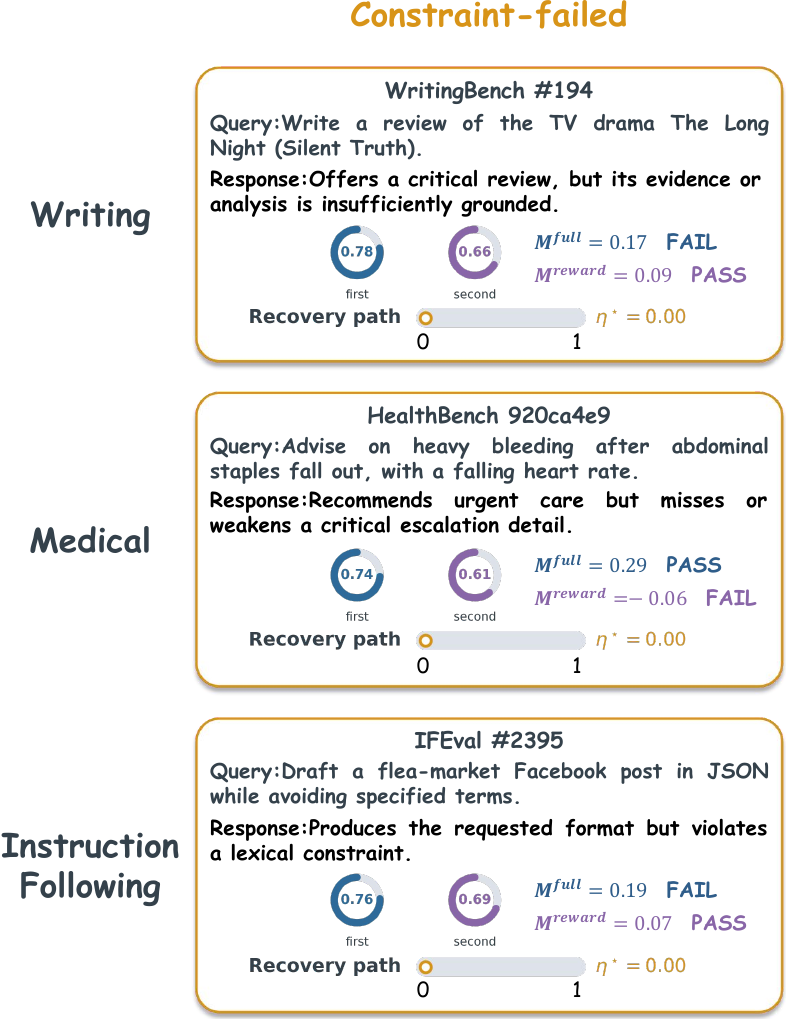}
    \caption{Representative rejected cases.
    Each case shows the anonymized query--response record, point and reference
    scores, full-geometry and reward-space projection checks, the recovery path
    from $\eta=0$ to $\eta=1$, the recovered training coefficient
    $\hat{\mathbf a}_x$, and the final decision.}
    \label{fig:representative_cases_2}
\end{figure*}

\begin{figure*}[htbp]
    \centering
    \includegraphics[width=\textwidth]{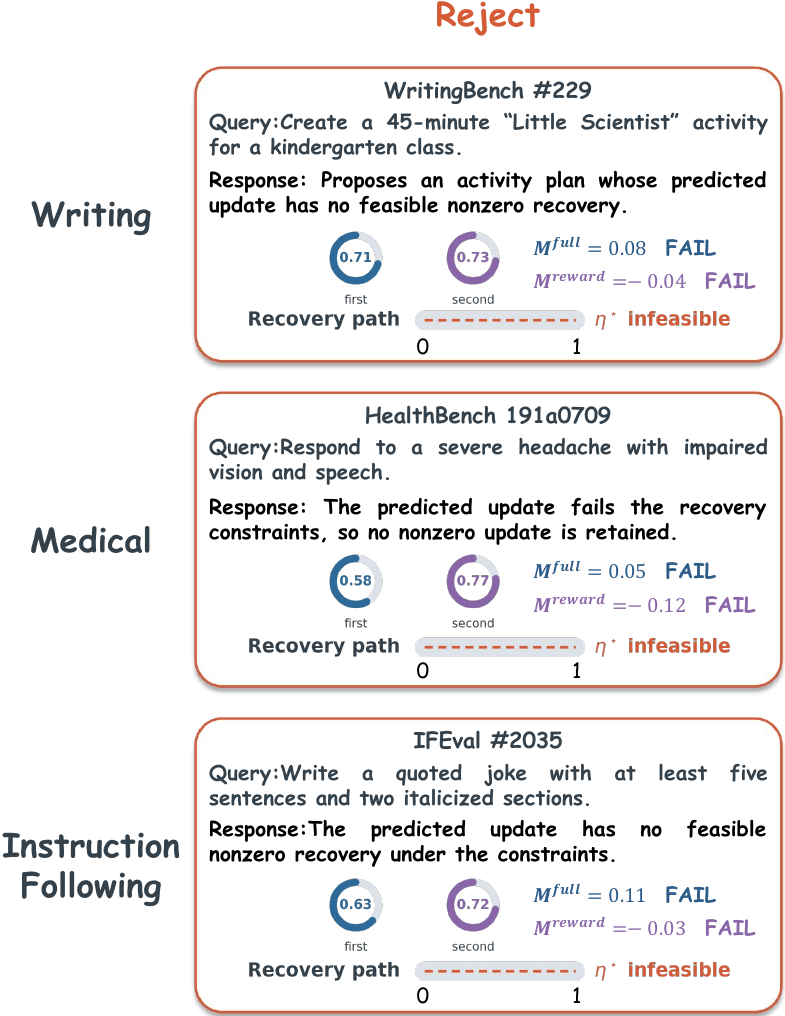}
    \caption{Representative recovery-infeasible cases.
    Each case shows the anonymized query--response record, point and reference
    scores, full-geometry and reward-space projection checks, the recovery path
    from $\eta=0$ to $\eta=1$, the recovered training coefficient
    $\hat{\mathbf a}_x$, and the final decision.}
    \label{fig:representative_cases_3}
\end{figure*}

Figures~\ref{fig:representative_cases_1}--\ref{fig:representative_cases_3} present three types of response groups: accepted, rejected, and recovery-infeasible, with three representative cases for each type. Each case reports the two scores, the corresponding training coefficients, the minimum projections in the policy-update space and reward space, the recovery ratio $\eta_x^\star$, and the final decision rationale. For the accepted cases, consider WritingBench \#4 as an example. Its two minimum projections are 0.84 and 0.81, satisfying $\Psi_x^{\mathrm{full}}\geq\tau_{\mathrm{sel}}$ and $\Psi_x^{\mathrm{reward}}>0$, with a maximum feasible recovery ratio of $\eta_x^\star=0.82$. The rejected cases all have nonzero candidate updates but fail at least one projection condition. For example, HealthBench 920ca4e9 has $M^{\mathrm{full}}=0.29$, satisfying the full-space condition, but $M^{\mathrm{reward}}=-0.06$, which fails the positive reward-space projection condition. Its recovery ratio is therefore $\eta^\star=0.00$. The recovery-infeasible cases have no nonzero feasible recovery ratio satisfying Eq.~\ref{eq:ggsp_recovery}. For example, IFEval \#2035 has $M^{\mathrm{full}}=0.11$ and $M^{\mathrm{reward}}=-0.03$, neither of which satisfies the corresponding condition, and is therefore marked as infeasible. The two scores do not vary monotonically with the decision outcome across the three types of cases. In contrast, the minimum projections and maximum feasible recovery ratio directly correspond to the screening conditions in Eq.~\ref{eq:ggsp_recovery}.

The decision differences across cases arise because the same score discrepancy induces different local updates under different response-level gradients. SetGRPO represents the joint uncertainty of the resulting training coefficients by $\mathcal C_x$, then combines it with the response-level gradients of the current policy to evaluate the minimum projections of candidate updates in the policy-update space and reward space and determine the maximum recovery ratio that can be written back for each response group. Thus, response groups satisfying the projection conditions retain nonzero training updates, whereas groups that fail these conditions do not write back an update.

%% file: sec/6_conclusion.tex
% !TEX root = ../neurips_2025.tex

% \vspace{-8pt}
\newpage
\section{Conclusion}
% 本文提出了 \textbf{SetGRPO}，用于处理 grader 随机性对 rubric-based GRPO 策略更新的影响。我们所设计的预实验表明，GRPO 的组内归一化会将其转化为相互关联的训练系数变化，当前策略的 response-level gradients 决定这些变化如何影响局部更新，而选择性执行则进一步改变被接受回答组的组成及其总体方向反转风险。基于这一观察，SetGRPO 在训练系数空间中构造联合不确定性集合，通过 JRSC 校准可能的系数变化，再利用 GGSR 结合当前策略的更新几何恢复可接受的训练信号，最后通过 SRGO 对接受更新的条件方向反转风险进行校准控制，同时保持原始 GRPO 的优化结构不变。
% 在三个领域的十个开放式任务基准上，SetGRPO 在Qwen-3.5与Llama-3.2模型簇上均取得所有对比训练方法中的最优宏平均性能。消融实验进一步支持了联合系数校准、策略局部更新和选择性风险控制对最终性能的贡献。

This paper proposed SetGRPO to address the effect of grader stochasticity on policy updates in rubric-based GRPO. Our pilot experiments showed that within-group normalization in GRPO converted grader stochasticity into correlated shifts in the training coefficients, while the response-level gradients of the current policy determined how these shifts affected local updates. Selective execution further changed the composition of the accepted response groups and their aggregate direction-reversal risk. Based on this observation, SetGRPO constructed a joint uncertainty set in training-coefficient space, calibrated possible coefficient shifts through JRSC, used GGSR to recover acceptable training signals in conjunction with the update geometry of the current policy, and finally calibrated and controlled the conditional direction-reversal risk of accepted updates through SRGO, while preserving the optimization structure of the original GRPO.
Across ten open-ended task benchmarks in three domains, SetGRPO achieved the best macro-average performance among all compared training methods on the Qwen-3.5 and Llama-3.2 model families.

\paragraph{Ethics Statement}

SetGRPO uses paired rubric-grader scores and response-level gradients from the
current policy to select and recover reinforcement-learning updates. These
components may introduce several risks: grader judgments may be stochastic,
biased, or subject to distribution shift; calibration data may contain
sensitive information; and the coefficient calibration and selective update
mechanisms may amplify artifacts present in the data or grader. The
finite-sample guarantee in this work concerns only the conditional
direction-reversal risk of accepted updates under the stated exchangeability
assumption. It does not guarantee privacy, fairness, factuality, or
harmlessness for individual responses.
The training, coverage-calibration, and risk-calibration splits are separated
by problem root, and downstream evaluation benchmarks are not used for
training, calibration, or checkpoint selection. Any deployment should apply
data minimization, appropriate authorization and access controls, and
anonymization where applicable. It should also include regular audits for
grader drift and subgroup disparities, as well as human review for
high-risk applications. Updates should be suspended or reviewed when the
exchangeability assumption, calibration conditions, or grader reliability are
not satisfied.

\paragraph{AI Assistance Statement}

During the preparation of this manuscript, the authors used generative AI
tools only to polish English writing, check grammar, and improve exposition.
Generative AI tools were not used to design the method, derive the theory,
conduct the experiments, or analyze the results. All AI-assisted text was
reviewed by the authors, who take responsibility for the final content of
this work.